\documentclass{article}

\usepackage{microtype}
\usepackage{graphicx}
\usepackage{subcaption}
\usepackage{booktabs} 

\usepackage{hyperref}

\usepackage[accepted]{arxiv}

\usepackage{amsmath}
\usepackage{amssymb}
\usepackage{mathtools}
\usepackage{amsthm}

\usepackage[capitalize,noabbrev]{cleveref}

\theoremstyle{plain}

\theoremstyle{definition}

\theoremstyle{remark}

\usepackage[textsize=tiny]{todonotes}

\newcommand{\method}{\textsc{VCPO}\xspace}

\icmltitlerunning{Stable Asynchrony: Variance-Controlled Off-Policy RL for LLMs}

\begin{document}

\twocolumn[
  \icmltitle{Stable Asynchrony: Variance-Controlled Off-Policy RL for LLMs}




  \begin{icmlauthorlist}
    \icmlauthor{Luke J. Huang}{mit}
    \icmlauthor{Zhuoyang Zhang}{mit}
    \icmlauthor{Qinghao Hu}{mit}
    \icmlauthor{Shang Yang}{mit}
    \icmlauthor{Song Han}{mit,nvidia}
  \end{icmlauthorlist}

  \icmlaffiliation{mit}{MIT}
  \icmlaffiliation{nvidia}{NVIDIA}

  \icmlcorrespondingauthor{Luke Huang}{lukh23@mit.edu}
  \icmlcorrespondingauthor{Song Han}{songhan@mit.edu}


  \vskip 0.3in
]



\printAffiliationsAndNotice{\icmlEqualContribution}

\begin{figure*}[!htb]
\vskip 0.1in
\centering
\includegraphics[width=1.0\linewidth]{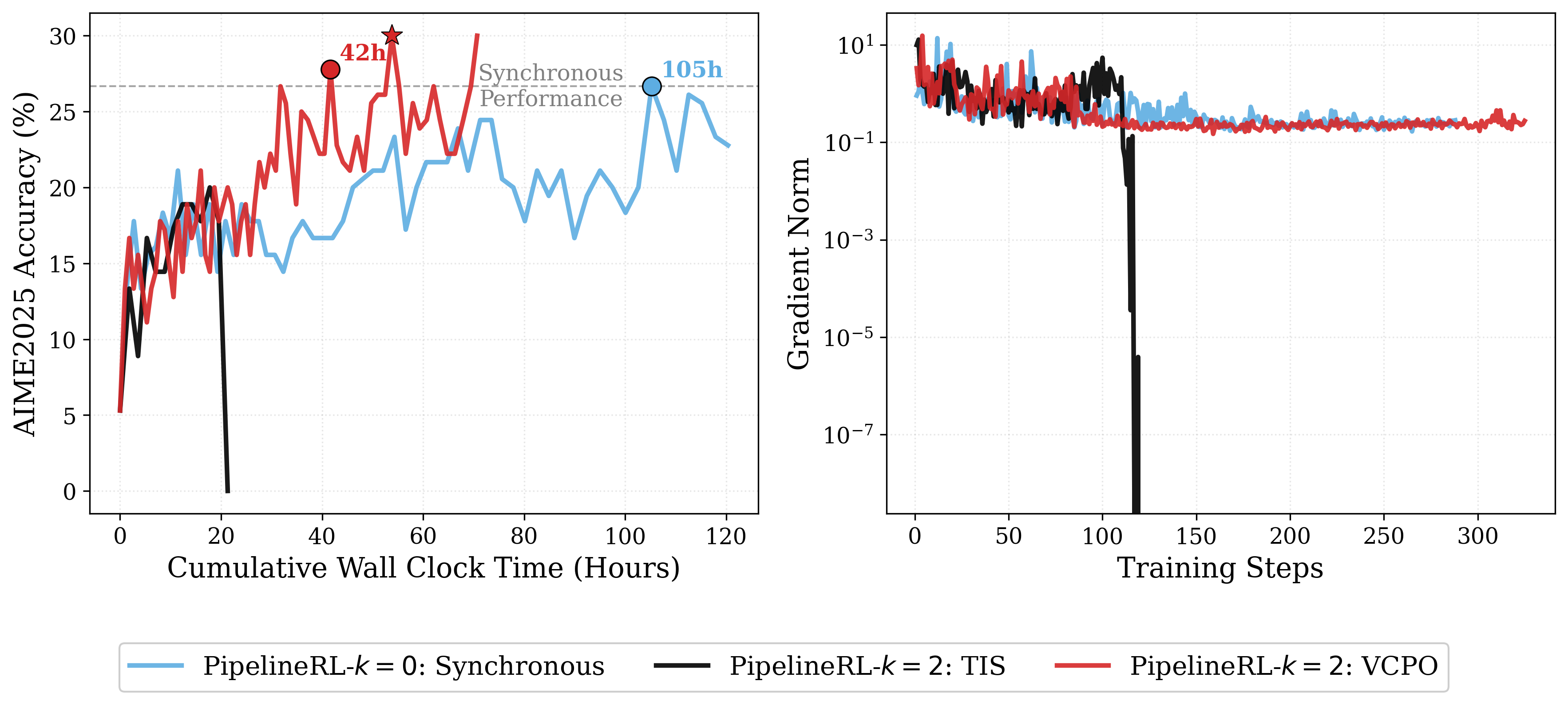}
\vskip -0.05in
\caption{
Long-context, tool-integrated multi-turn RL comparing synchronous training ($k{=}0$) to a two-step policy lag ($k{=}2$) with \method versus sequence-level truncated importance sampling (TIS).
Left: AIME-2025 validation accuracy vs.\ cumulative wall-clock time. \method matches the best synchronous accuracy $2.5\times$ faster ($\approx$42h vs.\ $\approx$105h) and continues to improve thereafter. Right: Gradient norm vs.\ training steps. The TIS run shows an instability characterized by a brief gradient-norm spike followed by rapid collapse.}
\label{fig:teaser}
\end{figure*}

\begin{abstract}
Asynchronous reinforcement learning has become increasingly central to scaling LLM post-training, delivering major throughput gains by decoupling rollout generation from policy updates. However, widely used policy-gradient objectives such as REINFORCE and GRPO suffer under high asynchrony: stale rollouts produce heavy-tailed importance weights, so a small number of trajectories dominate updates and the policy-gradient estimator becomes markedly \emph{higher variance}. Through systematic analysis on math, reasoning, and tool-use benchmarks, we find that this increasing variance is reliably predicted by collapsing effective sample size (ESS), which prior stabilization methods largely fail to address. Motivated by this diagnosis, we introduce \textbf{V}ariance \textbf{C}ontrolled \textbf{P}olicy \textbf{O}ptimization (\textbf{\method}), a method that (i) dynamically scales the learning rate with ESS to dampen unreliable updates and (ii) applies a closed-form minimum-variance baseline for off-policy settings, without a critic model and adding minimal overhead. Empirically, across math and general reasoning benchmarks, this enables robustly stable asynchronous training compared to previous stabilization and algorithmic methods, even in highly off-policy regimes (128 steps off-policy). In a long-horizon, tool-use task, \method matches synchronous performance while delivering a 2.5$\times$ speedup in training time. Code is available at: \url{https://github.com/mit-han-lab/vcpo}

\end{abstract}


\section{Introduction}
\label{sec:introduction}

Policy-gradient (PG) methods directly optimize a parameterized policy from sampled trajectories and have become the dominant approach for improving reasoning in large language models. Common objectives include Proximal Policy Optimization (PPO)~\cite{schulman2017ppo} as well as REINFORCE~\cite{williams1992reinforce} and its recent variants such as GRPO~\cite{shao2024deepseekmath}. 

These methods are typically implemented in an on-policy or near on-policy regime, where data generation and learning must occur sequentially after each other. This coupling becomes a major systems bottleneck for long-horizon tasks: the distribution of rollout lengths exhibit a long tail \cite{hu2025taminglongtail}, leading to pipeline bubbles and poor GPU utilization in synchronous RL training. 

To improve end-to-end training speed, asynchronous RL fully decouples rollout generation from learning via asynchronous, pipelined training. Prior works such as asynchronous RLHF~\cite{noukhovitch2024asyncrlhf} AReaL~\cite{fu2025areal}, LlamaRL~\cite{wu2025llamarl}, and PipelineRL~\cite{piche2025pipelinerl} have demonstrated up to 3$\times$ end-to-end speedups. However, this efficiency comes at the cost: training becomes strongly off-policy as the sampler generates trajectories with a lagging (stale) behavior policy. While importance sampling offers theoretically unbiased correction estimator, it fails to prevent learning degradation or even collapse in practice. Previous works have largely addressed this by suppressing outlier importance sampling ratios via masking, clipping, or other whitening and reshaping techniques, but in practice we find that these methods are insufficient for highly asynchronous training.

In contrast, we identify a deeper optimization failure mode: asynchronous RL training does not \textit{explicitly} account for the variance of the policy-gradient estimator itself. Across math, general reasoning, and tool-use benchmarks, we show that collapse is reliably predicted by a collapsing effective sample size (ESS)~\cite{kish1995surveysampling_reprint} of the importance-weighted gradient estimate, suggesting that collapse is preceded by increasing gradient variance. 

Motivated by this insight, we propose \textbf{V}ariance \textbf{C}ontrolled \textbf{P}olicy \textbf{O}ptimization (\method): a method with explicit variance-targeted controls that stabilize policy-gradient objectives for asynchronous RL training. \method consists of two complementary components: (i) an \emph{effective sample size guided step scaling} that down-weights unreliable updates and (ii) a \emph{closed-form minimum-variance off-policy reward baseline} based on per-trajectory gradient norms, avoiding training an auxiliary critic or value model. To make (ii) practical at scale, we present the first scalable implementation for \emph{exact} per-example gradient-norm computation with negligible runtime overhead, and leverage these statistics to build a low-overhead, gradient-aware baseline for variance reduction. 

Across highly asynchronous training regimes and across mathematical reasoning, general reasoning, and tool-use tasks, \method remains robust and effective from 1.5B to 7B models, outperforming a broad suite of baselines spanning masking/clipping stabilizers and algorithmic variants up to $128\times$ steps of off-policyness. When applied to long-context multi-turn training, \method reduces end-to-end training time by 2.5$\times$ while matching synchronous performance.

\section{Background}
\label{sec:background}

\subsection{Reinforcement Learning for LLMs}

We consider finetuning a pretrained language model policy $\pi_\theta$ to maximize a scalar reward $R(x,y)$ that scores a completion $y=(y_1,\dots,y_T)$ for a prompt $x$ (e.g., from a reward model or task-specific evaluator). 
Under an autoregressive policy, the likelihood of generating $y$ is
\[
\pi_\theta(y \mid x) \;=\; \prod_{t=1}^{T} \pi_\theta(y_t \mid x, y_{<t}).
\]
Let $\mathcal{D}$ denote a dataset (distribution) of prompts. RL optimizes the expected reward
\[
J(\theta) \;=\; \mathbb{E}_{x \sim \mathcal{D}, \,y \sim \pi_\theta(\cdot \mid x)}[R(x,y)].
\]

\textbf{On-Policy Gradient Estimator}. In On-Policy RL, we estimate $\nabla_\theta J(\theta)$ with sequences $y$ sampled from the policy $\pi_\theta$ itself. This leads to the REINFORCE gradient~\cite{williams1992reinforce}
\[
\nabla_\theta J(\theta)
=
\mathbb{E}_{x\sim\mathcal{D},\,y \sim \pi_\theta}\!\left[
R(x,y)\,\nabla_\theta \log \pi_\theta(y \mid x)
\right]
\]
\[
\nabla_\theta \log \pi_\theta(y \mid x) = \sum_{t=1}^{T} \nabla_\theta \log \pi_\theta(y_t \mid x, y_{<t})
\]
which motivates a surrogate loss of 
\begin{equation}
\mathcal{L}_{\text{on-policy}}(\theta)
=
-\mathbb{E}_{x\sim\mathcal{D},\, y\sim\pi_{\theta}(\cdot\mid x)}
\!\left[
R(x,y)\,\log \pi_\theta(y\mid x)
\right]
\label{eq:on_policy_loss}
\end{equation}
A standard variance reduction technique is to subtract a baseline that may depend on the state
$s_t = (x, y_{<t})$, but not on the current action/token $y_t$:
\[
\nabla_\theta J(\theta)
=
\mathbb{E}_{x,\,y \sim \pi_\theta}\!\left[
\sum_{t=1}^{T} A_t\, \nabla_\theta \log \pi_\theta(y_t \mid x, y_{<t})
\right],
\]
where an advantage-like term can be written as
\[
A_t \;=\; R(s_t, y_t) - b(s_t),
\qquad b(s_t) = b(x, y_{<t}).
\]
In actor--critic methods such as PPO, the baseline is a learned value function $b(s_t)=V_\phi(s_t)$ and advantages $A_t$ are computed via GAE. Critic-free methods such as GRPO instead use structured baselines (e.g., per-prompt group normalization), avoiding the overhead of training and serving a value model. In this paper, we focus on critic-free, REINFORCE-style updates, which are widely used for LLM reasoning post-training due to their lower resource requirements.

\textbf{Off-Policy Gradient Estimator}. In practical LLM RL systems, rollout generation and training are inevitably mismatched, causing training to be off-policy. This mismatch can arise even without explicit asynchronous training due to numerical issues~\cite{qi2025fp16mismatch,xi2026jetrl}, kernel and hardware issues~\cite{liu-li-2025-rl-collapse}, or unavoidable differences in inference and training implementations~\cite{he2025defeating_nondeterminism_llm_inference}. We therefore distinguish the \emph{sampler} policy $\mu$ used to generate trajectories from the current \emph{learner} policy $\pi_{\theta}$ being optimized. 

Importance Sampling (IS) corrects this mismatch. For a prompt $x$ and completion $y$ consisting of tokens $y_1, \cdots, y_T$, define the (sequence-level) IS ratio
\begin{equation}
w(x,y) \;\triangleq\; \frac{\pi_\theta(y\mid x)}{\mu(y\mid x)}  = \prod_{t=1}^T \frac{\pi_\theta(y_t \mid y_1^{t-1} x)}{\mu(y_t \mid y_1^{t-1} x)} 
\label{eq:is_weight}
\end{equation}
With importance sampling, Eq~\eqref{eq:on_policy_loss} can be expressed as an expectation relative to the \emph{sampler policy}, which leads to the unbiased off-policy surrogate loss
\begin{equation}
\begin{aligned}
\mathcal{L}_{\text{off-policy}}(\theta)
&=
-\mathbb{E}_{x\sim\mathcal{D},\,y\sim\mu(\cdot\mid x)}
\bigl[
w(x,y)\,A(x,y)\, \\
&\qquad\qquad\qquad\qquad
\log \pi_\theta(y\mid x)
\bigr].
\end{aligned}
\label{eq:is_loss}
\end{equation}
where $A(x,y)$ denotes an advantage-like signal (e.g., $R-b$ or token-level advantages).

\subsection{Why Effective Sample Size?}
\label{sec:background_ess}
In practice, the core challenge of importance sampling is not bias but \emph{finite-sample variance}. Known as the \textit{Curse of the Horizon}~\cite{liu2018breaking,liu2020understanding}, the product structure in~\eqref{eq:is_weight} makes $w(x,y)$ highly sensitive to small per-token probability shifts, so the resulting weights can become heavy-tailed and a few samples may dominate each update 

To quantify this weight degeneracy, given per-sample gradient contributions $g_i$ and importance weights $w_i$, define standardized weights
$\tilde w_i = w_i / \sum_{j=1}^B w_j$ and the corresponding weighted estimator
$\hat g = \sum_{i=1}^{B} \tilde w_i g_i$.
A common diagnostic is the \textbf{effective sample size} (ESS)~\cite{kong1992standardized,kong1994sequential}:
\[
\mathrm{ESS}
\;\triangleq\;
\frac{\left(\sum_{i=1}^{B} w_i\right)^2}{\sum_{i=1}^{B} w_i^2}
\;=\;
\frac{1}{\sum_{i=1}^{B} \tilde w_i^2}
\;\in\; [1,B].
\]
ESS measures how many samples effectively contribute to the weighted estimate: if weights are nearly uniform then $\mathrm{ESS}\approx B$, while if a few weights dominate then $\mathrm{ESS}\ll B$.
Moreover, under mild conditions (e.g., weak correlation between $g_i$ and $w_i$), we have
\[
\mathrm{Var}(\hat g)
\;\approx\;
\Big(\sum_{i=1}^{B} \tilde w_i^2\Big)\,\mathrm{Var}(g)
\;=\;
\frac{1}{\mathrm{ESS}}\,\mathrm{Var}(g),
\]
Intuitively, this means the variance of off-policy gradient estimates only matches the on-policy averaging rate with $B$ replaced by $\mathrm{ESS}$. Thus, when ESS collapses, gradient estimates become substantially higher variance, leading to high step-to-step volatility and KL. For this reason, practical RL systems and algorithms often apply clipping \cite{schulman2017ppo}, truncation \cite{espeholt2018impala}, or masking which limit the influence of extreme weights, at the cost of introducing bias relative to the exact IS correction.

\subsection{Prior RL stabilization methods}
While prior methods have been proposed to stabilize LLM RL in on-policy or near-on-policy settings, we discuss these works as they closely related to asynchronous training's highly off-policy issues. These approaches broadly fall into three categories: (1) masking/clipping to suppress outlier updates, (2) algorithmic changes to the estimator/objective, and (3) system-side changes that reduce sampler--learner mismatch.

\textbf{Masking / clipping mechanisms.} Similar to truncation and clipping mechanisms proposed by previous works~\cite{espeholt2018impala, schulman2017ppo}, recent work has propsed masking or truncating samples with outlier importance-weight ratios to prevent them from dominating updates.

Token or sequence-level \textsc{Truncated Importance Sampling} (TIS) constrains the IS ratio maximum. TIS has been used in large-scale asynchronous systems such as \textsc{LlamaRL}~\cite{wu2025llamarl} and \textsc{AReal}~\cite{fu2025areal} as well as in near on-policy RL to correct learner--sampler mismatch in modern LLM RL training stacks~\cite{yao2025offpolicyrlblog,yao2025rollouttrainingmismatch}

Closely related is token or sequence-level \textsc{Masked Importance Sampling} (MIS), which discard away updates with extreme IS ratios. \textsc{IcePop} proposed token-level MIS to improve stability in MoE RL~\cite{zhao2025icepop} while other works motivate sequence-level MIS due training--inference mismatch analyses~\cite{liu-li-2025-rl-collapse}. Notably, \textsc{DeepSeek-V3.2} uses masking sequences above a geometric-mean threshold and with negative advantages, illustrating a practical masking rule used at scale~\cite{deepseekai2025v32}.

Finally, \textsc{M2PO} iteratively drops token-level losses until a second-moment proxy of importance weights falls below a threshold~\cite{zheng2025m2po,zheng2025m2po_notion}.

\textbf{Algorithmic changes.} \textsc{GSPO} improves training stability in MoE RL, defining importance ratios as the geometric mean of the token-level IS ratios over the sequence and performs sequence-level clipping/optimization ~\cite{zheng2025gspo}.

\textsc{Optimal Token Baseline} (OTB) derives an on-policy variance-minimizing baseline and proposes a logit-based proxy (“energy”) for gradient-norms without extra backward passes~\cite{li2025otb}.

\textbf{System-side changes.} A complementary direction reduces mismatch via numerical and systems alignment. Switching from BF16 to FP16 can reduce train–inference numerical inconsistencies and improve stability~\cite{qi2025fp16mismatch}, while Rollout Routing Replay (R3) stabilizes MoE RL by replaying inference-time expert routing during training~\cite{ma2025r3}. These mitigations are orthogonal to variance-aware optimization and can be combined with our approach.

\subsection{Asynchronous RL}
In this paper, we adopt \textsc{PipelineRL-}$k$~\cite{piche2025pipelinerl} for our asynchronous training setup. At learner update \(t\), the trainer optimizes the current policy \(\pi_{\theta_t}\) using trajectories sampled by a (possibly stale) behavior policy \(\mu=\pi_{\theta_{t'}}\), where \(k\) bounds the \emph{policy lag} \(t-t'\) (with \(k=0\) recovering the synchronous/on-policy regime). This induces off-policy updates whose severity increases with \(k\). We also follow PipelineRL’s \emph{in-flight} weight-update scheme, where sampling continues while weights are being updated. As a result, a single trajectory may span multiple policy versions.

Throughout this paper, we use \emph{sequence-level} importance sampling to correct for this lag, which is the theoretically grounded importance sampling for sequence-level policy gradients, and optimize the importance-weighted surrogate objective in \eqref{eq:is_loss}.

\section{\method: Robust, Scalable Asynchronous RL}

We develop \method, a variance-controlled optimization framework for asynchronous RL. We first diagnose collapse through effective sample size (ESS), showing that dominated, high-variance updates precede KL and reward collapse.
We then derive a ESS-aware learning rate scaling rule and the variance-minimizing off-policy baseline (OPOB) for importance-weighted policy gradients.
Finally, we give a single-backward implementation and combine these components into \method.

\subsection{Why does Asynchronous RL collapse?}

To understand training collapse under asynchrony, we run controlled experiments with Qwen2.5-7B Base on MATH~\cite{hendrycks2021math}, a dataset of competition-style mathematics problems scored by exact-match final-answer correctness, and evaluate on 500 held-out problems (MATH-500). We log three signals over training: (i) the KL divergence between the rollout policy and the current learner policy (policy lag), (ii) reward / validation accuracy, and (iii) the \emph{ESS ratio}, i.e., the effective sample size computed from sequence-level importance weights and normalized by the minibatch size. 

\textbf{Baseline choice for diagnostics.}
To isolate the effect of policy lag, we adopt a strong masking/clipping baseline for these didactic collapse experiments. Based on the masking and clipping sweeps in Appendix~\ref{app:ablation_clipping_masking}, we find that \emph{sequence-level} truncated importance sampling (TIS) with a high threshold (\(c=8.0\)) delays collapse the longest among masking/clipping methods. We therefore use this setting as our default off-policy correction when diagnosing collapse. 

\textbf{Optimizer and KL regularization.} For all experiments we use AdamW with weight decay (see Appendix~\ref{sec:opt_hparams} for details). Following prior asynchronous~\cite{fu2025areal,wu2025llamarl} and synchronous~\cite{yu2025dapo,shah2026comedy} RL works, we do not incorporate explicit KL regularization loss as it constrains learning ability (corroborated by our own experiments in Appendix~\ref{app:ablation_kl}).

\begin{figure}[t]
\vskip 0.1in
\centering
\includegraphics[width=\linewidth]{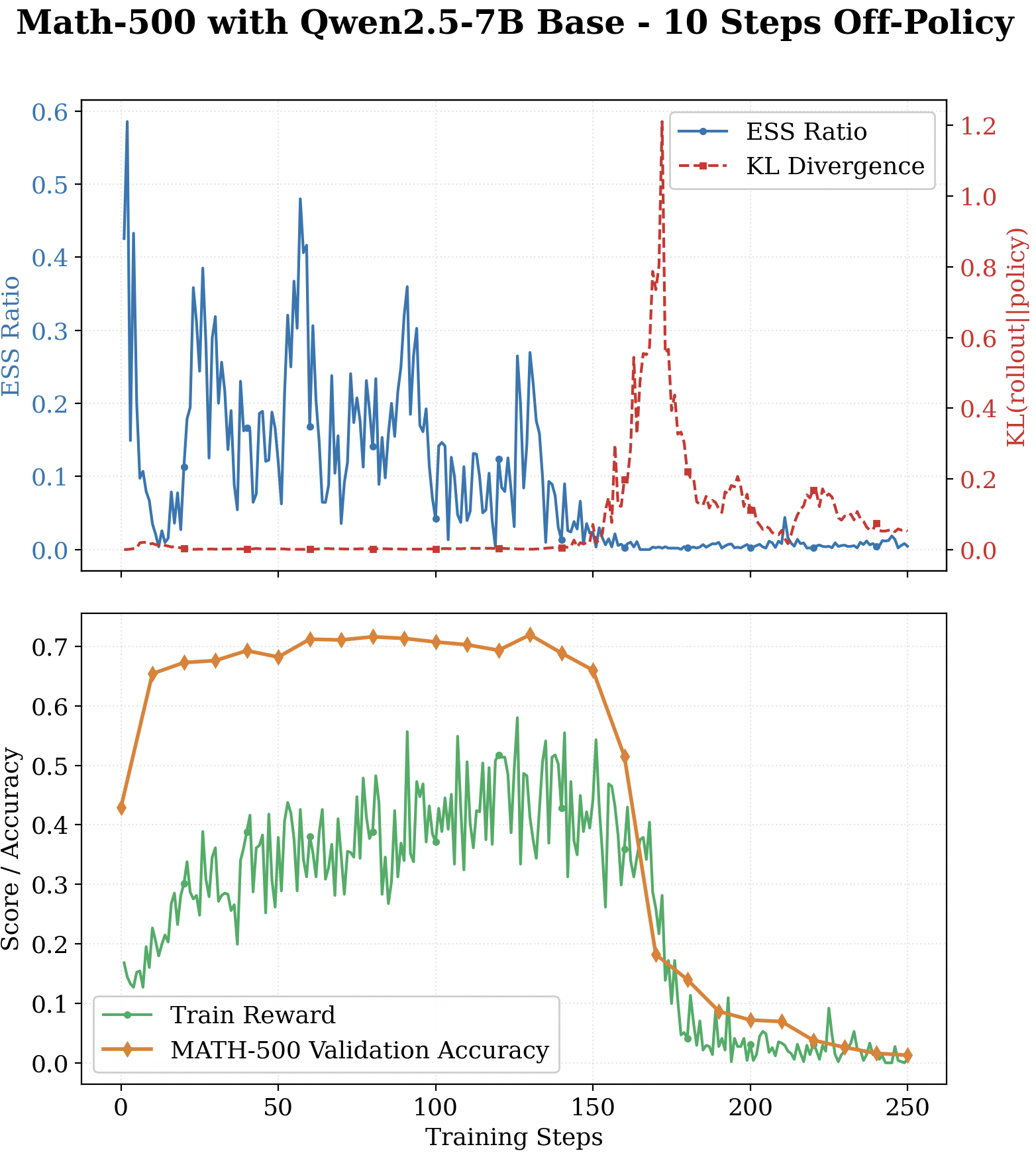}
\vskip -0.05in
\caption{\textbf{Sequence-Level TIS Collapse.}
 Qwen2.5-7B Base trained on MATH task with 10-step off-policy (\textsc{PipelineRL-}10). The ESS ratio first degrades and then collapses, leading to a spike in rollout--policy KL divergence and a sharp drop in both training reward and validation accuracy.
See Appendix~\ref{app:math_hyperparameters} for hyperparameters and training details.
}
\label{fig:didactic}
\end{figure}

\textbf{Training Collapse.} Figure~\ref{fig:didactic} shows a representative collapse under high staleness (10 steps off-policy). As the ESS ratio collapses, updates become dominated by a few trajectories, leading to a KL explosion and an abrupt drop in training reward and validation accuracy. This supports our hypothesis that \emph{high-variance, dominated updates} drives training collapse. For further evidence, see Appendix~\ref{app:ablation_collapse_examples} for examples on math and reasoning tasks.

\subsection{Effective Sample Size Guided Step-Size Scaling for Stable Updates} 

If asynchronous training fails due a subset of unreliable updates, a natural idea is to reduce the impact of those updates so that optimization remains robust under policy lag. A useful analogy comes from large-batch optimization: as batch size increases, gradient noise typically decreases~\cite{mccandlish2018largebatch}, allowing larger stable learning rates. While SGD often follows the \emph{linear scaling rule} (\(\eta \propto B\))~\cite{goyal2017largemini}, a widely used heuristic for adaptive optimizers such as Adam/AdamW is the \emph{square-root scaling law}: increasing minibatch size by a factor \(k\) permits scaling the learning rate by \(\sqrt{k}\) (equivalently, \(\eta \propto \sqrt{B}\))~\cite{krizhevsky2014oneweirdtrick,hoffer2017trainlonger,you2020lamb}.

In asynchronous off-policy RL, however, the \emph{nominal} batch size \(B\) can be misleading. As discussed in Section~\ref{sec:background_ess}, importance weights may be heavy-tailed, so a batch of size \(B\) can behave like a much smaller batch with only \(\mathrm{ESS}\) effectively independent samples. In this regime, \(\mathrm{ESS}\) is the relevant notion of effective batch size, and step-size selection should depend on \(\mathrm{ESS}\) rather than on \(B\).

Thus, motivated by the square-root scaling law \(\eta \propto \sqrt{B}\), we propose learning-rate scaling via an effective batch-size rule
\begin{equation}
\eta_{\text{eff}} \;\propto\; \sqrt{\rho_{\text{ess}}},
\qquad
\rho_{\text{ess}} \;\triangleq\;\frac{\mathrm{ESS}}{B}
\end{equation}
where we call $\rho_{\text{ess}}$ the \textbf{effective sample size ratio}. 

Because even on-policy/synchronous training can have $\rho_{\text{ess}}^{\text{on}}<1$ in practice (e.g., due to inevitable training-sampling differences and finite-sample effects), we introduce an \emph{empirical on-policy reference} $\rho^{{\text{on}}}_{\text{ess}} \triangleq \mathbb{E}[\rho_{\text{ess}}\mid \text{on-policy}]$ which is a \textbf{constant} estimated from 1 step of a on-policy run (or a running average of a few-steps)\footnote{Empirically, \(\rho_{\text{on}}\) is typically well-approximated by the first on-policy step.}. We then rescale the step size by the relative reliability of the batch:
\begin{equation}
\eta_{\text{eff}} \;=\; \eta \cdot \sqrt{\frac{\rho^{{\text{off}}}_{\text{ess}}}{\rho^{{\text{on}}}_{\text{ess}}}}
\label{eq:lr_ess_analogy}
\end{equation}
This scaling preserves the on-policy step scale when \(\rho_{\text{ess}}\approx\rho_{\text{on}}\), and automatically shrinks updates as \(\rho_{\text{ess}}\) collapses.

Thus when training is nearly on-policy and \(\mathrm{ESS}\approx B\) and \(\eta_{\text{eff}}\approx \eta\); when a few samples dominate, \(\mathrm{ESS}\ll B\) and \(\eta_{\text{eff}}\) shrinks like \(\sqrt{\mathrm{ESS}/B}\), damping the unstable, high-variance updates that coincide with KL/gradient spikes.

\subsection{Minimum Variance Baselines for Off-Policy RL}

Subtracting a baseline from the reward (or advantage) leaves the expected policy gradient unchanged but can substantially reduce variance. This control-variate view dates to reinforcement-comparison methods~\cite{dayan1991reinforcementcomparison} and was formalized for policy gradients~\cite{greensmith2004variance}, with optimal baselines derived in the on-policy setting~\cite{weaver2013optimalbaseline}.

In more recent work, systems often use simple group baselines (e.g., per-prompt mean reward) for efficiency~\cite{li2023remax,shao2024deepseekmath}, while recent work studies heuristic or approximate optimal baselines on-policy~\cite{hao2025opo,li2025otb}. In highly asynchronous training, however, importance weighting changes the variance structure. We thus derive the variance-minimizing baseline for an \emph{off-policy, importance-weighted} policy-gradient estimator.

\textbf{Setup.} Consider the off-policy gradient estimator with a scalar baseline $b$:
\begin{equation}
\hat{G}(b)
\;=\;
\frac{1}{B}\sum_{i=1}^B w_i (R_i - b)\, \nabla_\theta \log \pi_\theta(\tau_i)
\label{eq:offpolicy_grad_baseline}
\end{equation}
where each sample $i$ contribute a score-gradient vector $g_i = \nabla_\theta \log \pi_\theta(\tau_i)$,  a scalar reward/advantage $R_i$, and an importance ratio $w_i$.

\textbf{Optimal Off-policy Baseline.} As we derive in Appendix~\ref{app:offpolicy_baseline}), minimizing $\mathrm{Var}(\hat{G}(b))$ over scalar $b$ yields a closed-form solution for the off-policy optimal baseline (OPOB)
\begin{equation}
b_{\text{OPOB}}^\star
\;=\;
\frac{\sum_{i=1}^B w_i^2 \,\|g_i\|^2\, R_i}{\sum_{i=1}^B w_i^2 \,\|g_i\|^2}.
\label{eq:offpolicy_opt_baseline}
\end{equation}
Compared to common group baselines (e.g., $b=\frac{1}{B}\sum_i R_i$), Eq.~\eqref{eq:offpolicy_opt_baseline} shows that variance-optimal baselining in the off-policy regime depends on \emph{both}
(i) the importance weights $w_i$ and
(ii) the gradient magnitudes $\|g_i\|^2$.
Intuitively, samples that are both highly upweighted off-policy and induce large parameter changes dominate update variance and should therefore dominate the baseline.

\textbf{Connections to prior baselines.} Equation~\eqref{eq:offpolicy_opt_baseline} can recover many familiar special cases.
When training is on-policy ($w_i = 1$), it reduces to the classic gradient-norm-weighted optimal baseline~\cite{greensmith2004variance,weaver2013optimalbaseline}.
If, additionally, $\|g_i\|$ is approximately constant within a group, then $b^\star$ reduces to the group-mean reward baseline commonly used in practice. Our derivation shows that in the off-policy regime, \emph{both} importance weights and gradient magnitudes are required for variance-optimal baselining.

\subsection{Efficient Baseline-Aware Gradient Computation}
\label{sec:efficient_grad_norm}

A naive gradient-aware baseline requires two backward passes: one to compute per-sample gradient norms and a second to apply the resulting baseline, roughly doubling backward compute time at scale. We instead implement it with a single backward pass, adding minimal wall-clock and memory overhead.

\textbf{Key observation: linearity of the baseline term.} Note that expanding the equation in \eqref{eq:offpolicy_grad_baseline} yields
\begin{equation}
\begin{aligned}
\hat{G}(b)
&=
\frac{1}{B}\underbrace{\sum_{i=1}^B w_i R_i g_i}_{\text{reward-weighted term}}
-
b\cdot
\frac{1}{B}\underbrace{\sum_{i=1}^B w_i g_i}_{\text{score term}} \\
&=
\frac{1}{B}\bigl(G_R - b\,G_S\bigr).
\end{aligned}
\label{eq:offpolicy_grad_decomp}
\end{equation}
where $G_R \triangleq \sum_{i=1}^B w_i R_i g_i$ and $G_S  \triangleq \sum_{i=1}^B w_i g_i$. Thus, for any chosen $b$, the final gradient can be formed by combining two aggregated quantities $G_R$ and $G_S$.

\textbf{Score-gradient reweighting into two buffers.} We take advantage of this linearity by computing per-sample score gradients $g_i$ once and accumulate them into two gradient buffers:
(i) $G_R$ a buffer that accumulates the reward/advantage-weighted score gradients $R_i w_i g_i$, and
(ii) $G_S$ a score buffer that accumulates score gradients $w_i g_i$. Afterwards, we form the final update by a simple linear combination of these buffers following Eq. \ref{eq:offpolicy_grad_decomp}.
This avoids any additional backward pass to incorporate the baseline term.

\begin{algorithm}[t]
\caption{Single-backward accumulation with Off-Policy Optimal Baseline baseline}
\label{alg:single_backward_opob}
\begin{algorithmic}[1]
\STATE \textbf{Input:} Minibatch $\{(x_i,\tau_i,R_i,w_i)\}_{i=1}^B$, learner policy $\pi_\theta$
\STATE \textbf{Output:} Baseline-aware gradient estimate $\hat{G}(b^\star_{\text{OPOB}})$ and scalar baseline $b^\star_{\text{OPOB}}$
\STATE Initialize gradient buffers $G_R \leftarrow 0$, $G_S \leftarrow 0$
\STATE Initialize baseline numerators $N \leftarrow 0$, $D \leftarrow 0$
\FOR{$i=1$ \text{to} $B$}
    \STATE $\ell_i \leftarrow \log \pi_\theta(\tau_i \mid x_i)$
    \STATE $g_i \leftarrow \nabla_\theta \ell_i$ \hfill(\emph{backprop})
    \STATE $s_i \leftarrow \|g_i\|_2^2$
    \STATE $G_R \leftarrow G_R + (w_i R_i)\, g_i$
    \STATE $G_S \leftarrow G_S + w_i\, g_i$
    \STATE $N \leftarrow N + w_i^2\, s_i\, R_i$
    \STATE $D \leftarrow D + w_i^2\, s_i$
\ENDFOR
\STATE $b^\star_{\text{OPOB}} \leftarrow \mathrm{stopgrad}\!\left(\frac{N}{D+\varepsilon}\right)$
\STATE $\hat{G}(b^\star_{\text{OPOB}}) \leftarrow \frac{1}{B}\left(G_R - b^\star_{\text{OPOB}}\, G_S\right)$
\STATE \textbf{return} $\hat{G}(b^\star_{\text{OPOB}}),\, b^\star_{\text{OPOB}}$
\end{algorithmic}
\end{algorithm}

\begin{figure}[ht]
  \vskip 0.2in
  \begin{center}
    \centerline{\includegraphics[width=0.95\columnwidth]{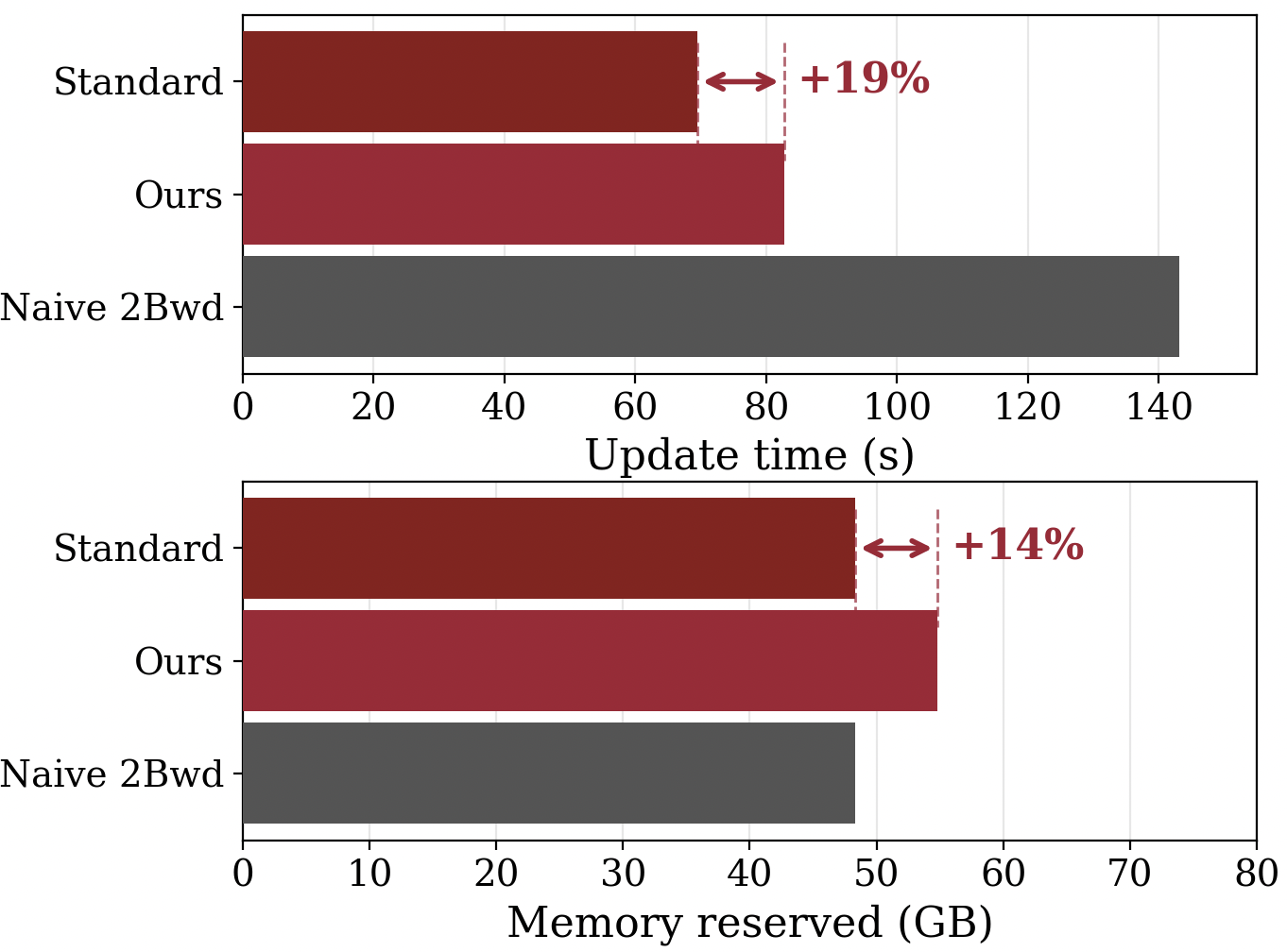}}
    \caption{ Compute (left) and memory (right) overhead of baseline-aware updates for Qwen2.5-7B on 4$\times$H100 GPUs (TP=4) with a sequence length of 8192 tokens. 
}
\label{fig:gradient-aware-backward}
  \end{center}
\end{figure}

\textbf{Implementation.} We implement Algorithm~\ref{alg:single_backward_opob} in \texttt{verl} using the Megatron-LM backend so that we can efficiently obtain gradient norm statistics. As can be seen in Figure~\ref{fig:gradient-aware-backward}, our method reduces the overhead of the naive method from 100\% to a 19\% step time overhead while only modestly increase memory usage by 14\%. See Appendix \ref{app:opob_implementation} for more implementation details.

\subsection{\method: Variance Controlled Policy Optimization}

As noted in Appendix~\ref{app:ablation_clipping_masking}, we find that \emph{sequence-level} truncated importance sampling (TIS) with a higher threshold (\(c=8.0\)) delays collapse the longest among all current stabilization methods. We thus combine sequence-level TIS with our effective-sample-size step-size scaling and our off-policy optimal baseline. This leads to the following surrogate loss, whose gradient equals our baseline-aware, truncated IS policy-gradient estimator
\begin{equation}
\begin{aligned}
\mathcal{L}_{\method}(\theta)
&=
-\mathbb{E}_{x\sim\mathcal{D},\,y\sim\mu(\cdot\mid x)}
\\
&\quad\Bigl[
w^{\textsc{TIS}}(x,y)\,(R(x,y)-b_{\text{OPOB}}^\star)\,\log \pi_\theta(y\mid x)
\Bigr]
\\
&\approx
-\frac{1}{B}\sum_{i=1}^{B} w^{\textsc{TIS}}_i\,(R_i-b_{\text{OPOB}}^\star)\,\log \pi_{\theta}(y_i\mid x_i),
\end{aligned}
\label{eq:VCPO_loss}
\end{equation}
where the sequence-level truncated importance weight is
\[
w^{\textsc{TIS}}(x,y)
\;=\;
\min\!\left(
\mathrm{sg}\left[\frac{\pi_\theta(y\mid x)}{\mu(y\mid x)}\right],\;
\;c
\right)\]
We then apply AdamW using the gradient \(\nabla_\theta \mathcal{L}_{\method}(\theta)\), but scale learning rate \(\eta\) with an ESS-scaled step size
\[
\eta_{\text{eff}} \;=\; \eta \cdot \sqrt{\frac{\rho^{{\text{off}}}_{\text{ess}}}{\rho^{{\text{on}}}_{\text{ess}}}}
\]
such each training step size scales with \(\mathrm{ESS}\) accordingly. Note although we adopt TIS, we still use the \emph{unclipped IS ratios} to calculate ESS.

\section{Empirical Evaluation}

\begin{figure}[t]
\vskip 0.1in
  \centering
  \includegraphics[width=1.0\linewidth]{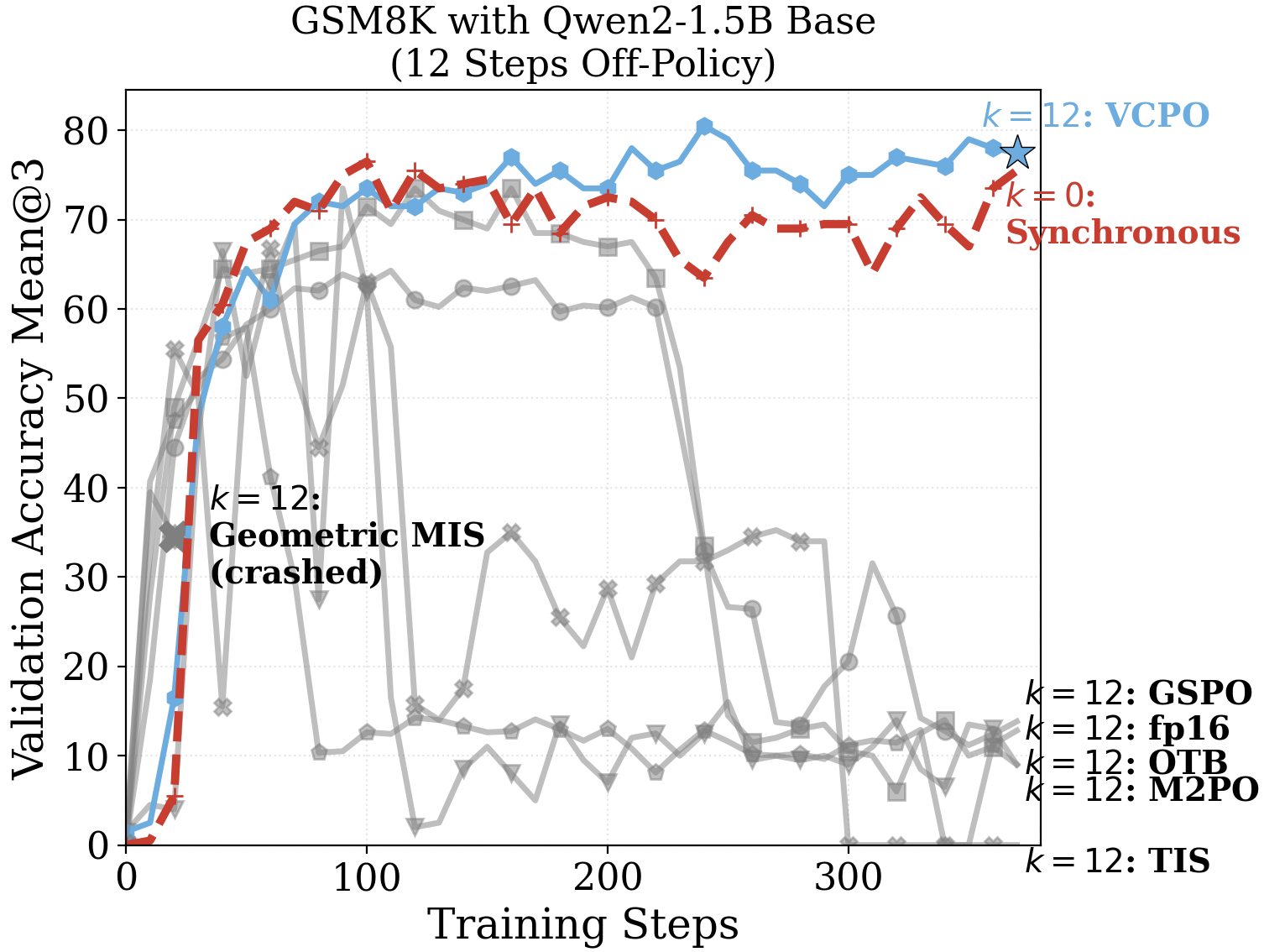}
\vskip -0.05in
\caption{
GSM8K with Qwen2-1.5B under \textsc{PipelineRL-}12 (high policy lag).
Most baselines lead to training collapse (or crash, e.g. Geometric MIS masks all sequences and has no loss), while \method remains stable throughout training and matches synchronous performance. Training details and hyperparameters can be found in Appendix~\ref{app:gsm8k_hyperparameters}.}
\label{fig:gsm8k_results}
\end{figure}



\begin{figure*}[t]
\vskip 0.1in
\centering
\includegraphics[width=1.0\linewidth]{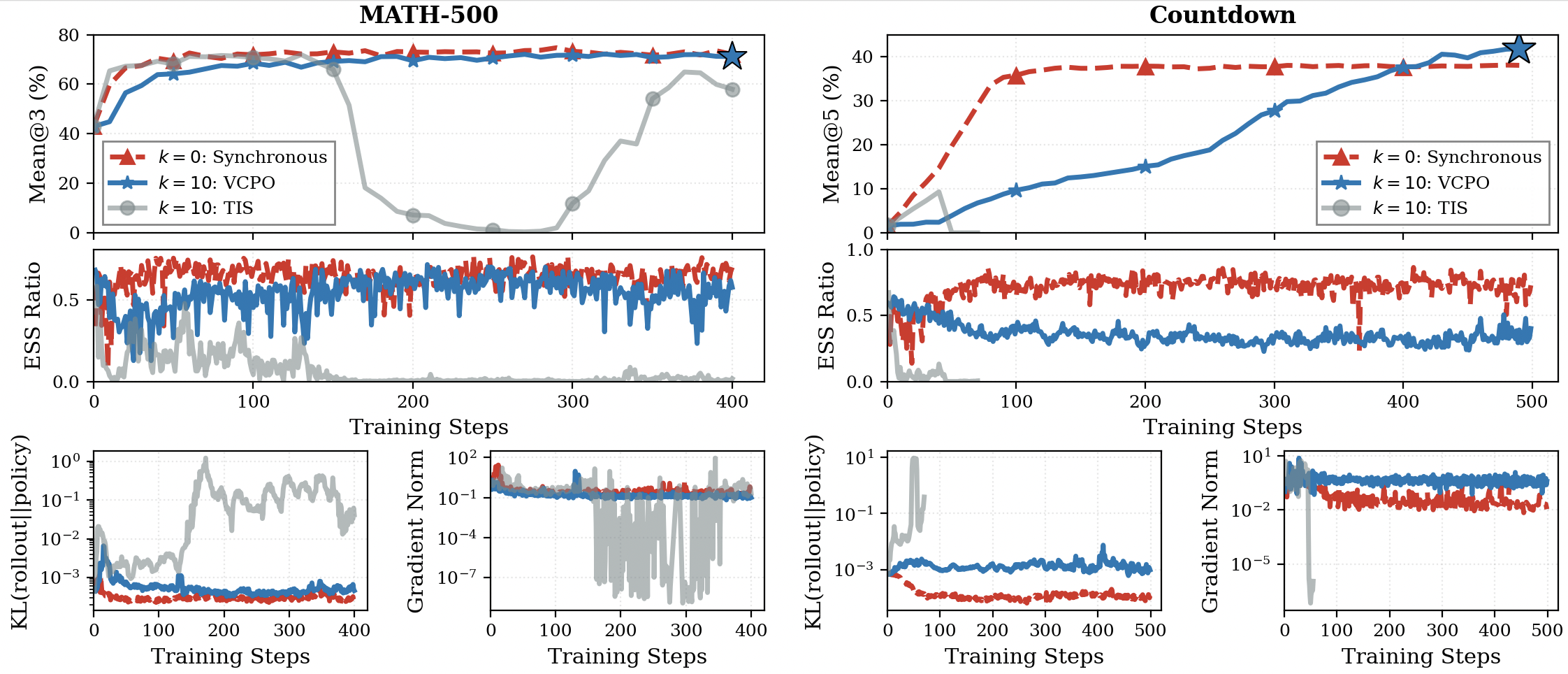}
\vskip -0.05in
\caption{
Qwen2.5-7B under \textsc{PipelineRL-}10 (10 Steps Off-Policy) on Countdown and MATH-500. Across both tasks, sequence level truncated importance sampling (TIS) suffers ESS collapse followed by KL/gradient instability and degraded accuracy, whereas \method maintains healthy ESS and stable updates, reaching synchronous performance. Training details and hyperparameters are provided in Appendix~\ref{app:math_hyperparameters}.
}
\label{fig:math_countdown_results}

\end{figure*}

\subsection{Tasks}

We evaluate \method across three representative RL post-training pipelines for LLM reasoning and agentic tool use in highly asynchronous regimes. Across all three domains, \method remains robust to substantial learner--sampler mismatch, improving throughput while preserving strong learning performance

\begin{enumerate}[topsep=2pt, itemsep=1pt, parsep=0pt, partopsep=0pt]
    \item \textbf{Mathematical problem solving.}
    We evaluate on GSM8K~\cite{cobbe2021gsm8k}, a benchmark of grade-school math word problems, and on MATH~\cite{hendrycks2021math}, a dataset of competition-style mathematics problems spanning multiple subjects and difficulty levels. For both benchmarks, we use a binary reward based on exact match of the final numeric answer and train on the official training split while evaluating on the official validation split.

    \item \textbf{General reasoning (verifiable).}
    We construct a Countdown-style arithmetic reasoning dataset using Reasoning Gym~\cite{stojanovski2025reasoninggym}, where rewards are verifiable by deterministic checking of the final answer.
    We sample 9{,}000 problems for training and hold out 1{,}000 problems for validation, following prior setups for verifiable-reward reasoning tasks~\cite{liu2025prorl}.

    \item \textbf{Long horizon tool-integrated reasoning.}
    We evaluate multi-turn tool use in the SimpleTIR setting~\cite{xue2025simpletir}, where the model must interleave reasoning with external tool calls. We train using the DAPO dataset~\cite{yu2025dapo} and evaluate on a held-out exam-style benchmark (AIME2025). This setting requires long rollouts (reasoning and multi-turn interactions with tool calls), so training is often dominated by rollout generation. As a result, strictly on-policy training underutilizes compute: the gradient updates must wait for sampling to complete, and this bottleneck worsens at scale and in sparse-reward domains that require more exploration.
\end{enumerate}

\subsection{Metrics and Baselines}

\textbf{On-Policy Training as Oracle Baseline}
To disentangle task-learning dynamics from instabilities introduced by asynchrony, we report a fully synchronous (\(k=0\)) RL run as an oracle reference (matched training steps and hyperparameters). 

\textbf{Other Baselines.} We compare against representative prior approaches spanning (i) masking/clipping-based stabilization (e.g., M2PO, MIS, TIS), (ii) objective-level modifications such as GSPO and OTB, and (iii) numerical/systems mitigations (e.g. using fp16 precision). Although (ii) and (iii) were not designed specifically for highly asynchronous RL, we include them for completeness because they are commonly used to stabilize on-policy LLM-RL training. In this paper, we train dense models, and so MoE specific methods such as Rollout Routing Replay are not considered.

\subsection{\method Delivers Synchronous Performance with Higher Throughput}
Across mathematical reasoning, general reasoning, and tool-use tasks, \method matches the final performance of fully synchronous RL even under high asynchrony. On GSM8K, Figures~\ref{fig:gsm8k_results} show that \method enables stable training while several baselines collapse or crash in this regime. For the Countdown and MATH-500 tasks, Figure~\ref{fig:math_countdown_results} shows that \method attains the same peak performance as synchronous RL with higher end-to-end throughput enabled by asynchronous training. Table \ref{tab:throughput_step_split} summarizes the end-to-end wall-clock gains of up to $2.5\times$.
Furthermore, component ablations show both ESS based learning rate scaling and the off-policy optimal baseline improve robustness, and combining them yields the best stability and accuracy (Appendix~\ref{app:ablation_component}).

\begin{table}[t]
\centering
\small
\setlength{\tabcolsep}{5pt}          
\renewcommand{\arraystretch}{1.08}    
\renewcommand{\arrayrulewidth}{0.8pt} 

\caption{End-to-end training times and validation accuracy for synchronous vs.\ asynchronous training (lag $k$).}
\label{tab:throughput_step_split}

\begin{tabular}{l|c|c|c}
\specialrule{1.2pt}{0pt}{0pt}
\textbf{Method} &
\shortstack[c]{Countdown $\uparrow$} &
\textbf{Steps} &
\textbf{GPU hours $\downarrow$} \\
\specialrule{0.8pt}{0pt}{0pt}
\specialrule{0.6pt}{0pt}{0pt}

Base &
\shortstack[c]{1.6\%} &
-- &
-- \\

Sync ($k{=}0$) &
\shortstack[c]{38.4\%} &
400 &
143.2 \\

VCPO + Async ($k{=}10$) &
\shortstack[c]{41.9\%} &
400 &
\textbf{89.6} \\
\specialrule{0.8pt}{0pt}{0pt}
\end{tabular}

\vspace{3pt}

\begin{tabular}{l|c|c|c}
\specialrule{0.8pt}{0pt}{0pt}
\textbf{Method} &
\shortstack[c]{MATH-500 $\uparrow$} &
\textbf{Steps} &
\textbf{GPU hours $\downarrow$} \\
\specialrule{0.8pt}{0pt}{0pt}
\specialrule{0.6pt}{0pt}{0pt}

Base &
\shortstack[c]{40.2\%} &
-- &
-- \\

Sync ($k{=}0$) &
\shortstack[c]{72.0\%} &
400 &
134.4 \\

VCPO + Async ($k{=}10$) &
\shortstack[c]{71.6\%} &
400 &
\textbf{92.8} \\
\specialrule{0.8pt}{0pt}{0pt}
\end{tabular}

\vspace{3pt}
\begin{tabular}{l|c|c|c}
\specialrule{0.8pt}{0pt}{0pt}
\textbf{Method} &
\shortstack[c]{AIME 2025 $\uparrow$} &
\textbf{Steps} &
\textbf{GPU hours $\downarrow$} \\
\specialrule{0.8pt}{0pt}{0pt}
\specialrule{0.6pt}{0pt}{0pt}

Base &
\shortstack[c]{5.3\%} &
-- &
-- \\

Sync ($k{=}0$) &
\shortstack[c]{26.7\%} &
300 &
420.2 \\

VCPO + Async ($k{=}2$) &
\shortstack[c]{27.8\%} &
220 &
\textbf{168.9} \\
\specialrule{1.2pt}{0pt}{0pt}
\end{tabular}
\end{table}
\vspace{-1pt}
\subsection{\method Provides State-of-the-Art Stability Under High Asynchrony}

\textbf{Enabling Longer Stable Training.}
Figure~\ref{fig:gsm8k_results} summarizes our highly asynchronous training results (12 steps off-policy) on GSM8K with Qwen2-1.5B. While the synchronous baseline trains smoothly, most existing stabilization approaches fail under this level of policy lag. 

In contrast, \method consistently avoids collapse. Across baselines: (i) ratio-based clipping (\textsc{TIS}) and masking (\textsc{MIS}) eventually collapse; (ii) algorithmic variants such as \textsc{GSPO} and \textsc{OTB} do not reliably prevent collapse; and (iii) systems-side mitigations (e.g., FP16) delay collapse but do not address the underlying instability mechanism. 

See Appendix~\ref{app:ablation} for additional comparisons and ablations, including systematic sweeps over \textsc{TIS}/\textsc{MIS} hyperparameters, algorithmic and systems variants (\textsc{GSPO}, \textsc{OTB}, learning-rate tuning), and tasks.

\textbf{Avoiding Effective Sample Size Collapse and Training Degradation.}
As shown in Figure~\ref{fig:math_countdown_results}, in both general mathematical (MATH-500) and reasoning tasks (Countdown), naively training under high asynchrony leads to ESS-ratio collapse followed by gradient instability and performance degradation. However, \method remains stable, maintaining a high ESS ratio, stable gradient norms, and KL levels throughout training.

\begin{figure}[!htb]
\vskip 0.1in
\centering
\includegraphics[width=1.0\linewidth]{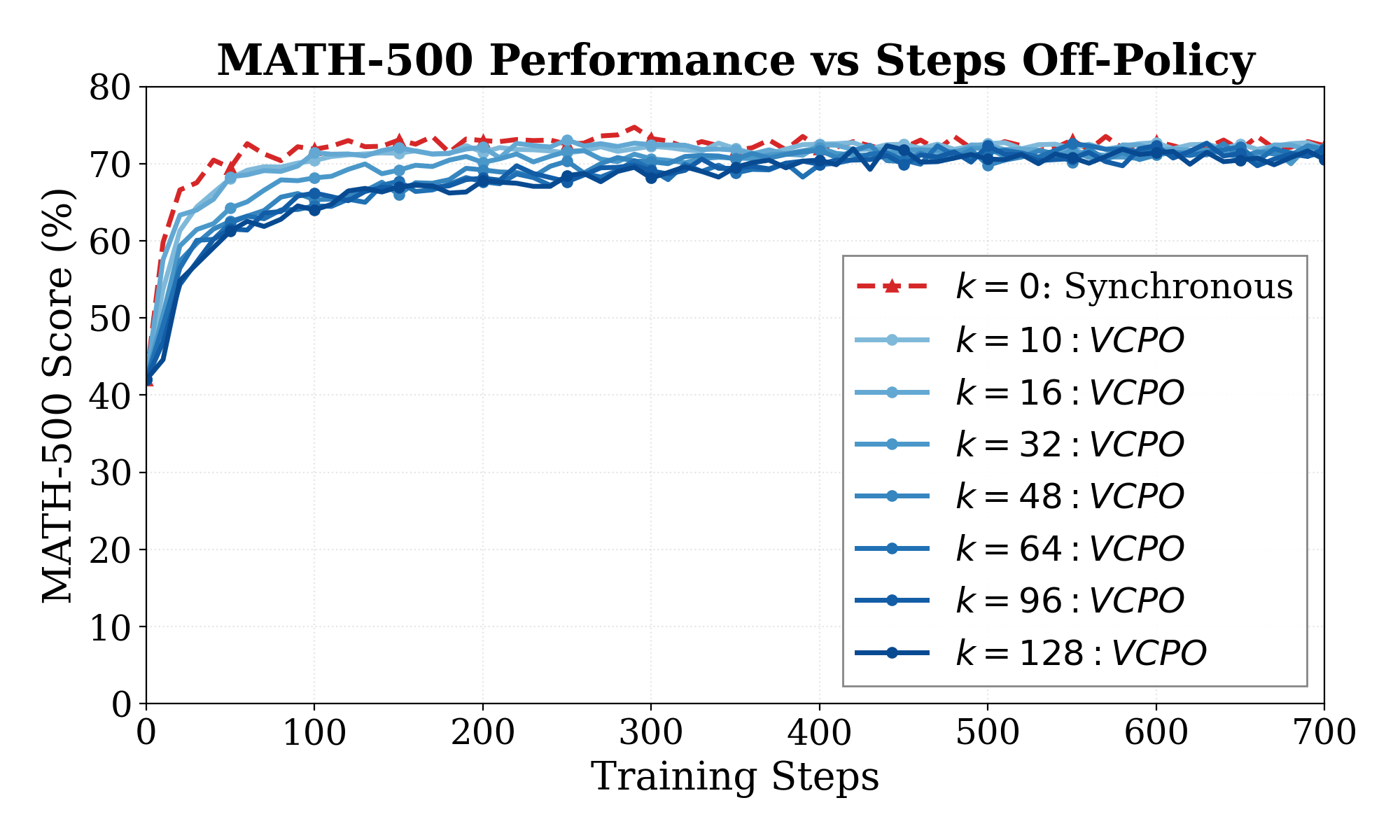}
\vskip -0.05in
\caption{\textbf{Stable training under extreme policy lag}: Qwen2.5-7B trained with PipelineRL at policy lag $k$ (steps off-policy). VCPO is stable up to at least $k{=}128$ steps off-policy and matches the synchronous baseline ($k{=}0$).}
\label{fig:high_staleness}
\end{figure}

\textbf{Scaling to extreme staleness.}
While the end-to-end training speedups from asynchronous are fully saturated for $k=10$ steps off-policy, for completeness we stress test stability as policy lag increases. In Figure~\ref{fig:high_staleness}, we see 
\method training remains stable up to at least $k{=}128$ steps off-policy while matching synchronous training performance ($k{=}0$).

\subsection{\method Enables Fast and Stable Long-Horizon RL}
We further evaluate \method in a long-context, tool-integrated multi-turn setting following SimpleTIR~\cite{xue2025simpletir}, where rollouts are substantially longer and training is known to be more brittle~\cite{li2025otb}. We train with minibatch size 128 and 16 responses per prompt, using a maximum completion length of 12K tokens and up to 5 tool calls per trajectory. We train on the DAPO~\cite{yu2025dapo} dataset and evaluate on AIME 2025, see Appendix~\ref{app:multiturn_hyperparameters} for more details.

Due to longer trajectories and prevalence of out-of-distribution tool calls, we use a lower level of asynchrony (2 steps off-policy). Figure~\ref{fig:teaser} shows that unlike sequence-level TIS which collapses, \method improves steadily in both train reward and AIME2025 accuracy, reducing end-to-end training time--a by $2.5\times$ compared the synchronous baseline. This demonstrates that our variance-control mechanisms transfer to long-horizon, multi-turn post-training.

\section{Discussion}

In this work, we introduce \method, a drop-in method for robust, off-policy asynchronous RL LLM post-training. We identify a consistent failure mode in highly asynchronous regimes-- an effective sample size collapse that amplifies policy gradient variance--and building on this diagnosis, \method stabilizes training with two main contributions 
(i) \textit{effective-sample-size–guided step-size rescaling} to automatically damp unreliable updates, and (ii) a \textit{closed-form minimum-variance baseline} for importance-weighted gradients that avoids training an auxiliary value model and adds minimal overhead. Across mathematical reasoning, general reasoning, and long-context multi-turn tool use, \method improves robustness under substantial learner–sampler mismatch while preserving the throughput benefits of asynchronous rollouts, highlighting explicit variance control as a key ingredient for reliable asynchronous RL at scale.

\textbf{Limitations and future work.} We study dense Transformer policies under standard precision and a fixed set of asynchronous pipeline designs. Future work should evaluate \method on MoE models (where routing can amplify mismatch), under more aggressive quantization (e.g., FP8), and in substantially longer-horizon agentic settings (e.g., search/planning with sparse rewards), where additional exploration and memory mechanisms may be needed alongside variance control.

\newpage


\section*{Acknowledgements}
We thank MIT-IBM Watson AI Lab, Amazon, Dell, and National Science Foundation for supporting this research. We thank NVIDIA for donating the DGX server.


\bibliography{example_paper}

@article{williams1992reinforce,
  title   = {Simple statistical gradient-following algorithms for connectionist reinforcement learning},
  author  = {Williams, Ronald J.},
  journal = {Machine Learning},
  volume  = {8},
  pages   = {229--256},
  year    = {1992},
  doi     = {10.1007/BF00992696},
  url     = {https://link.springer.com/article/10.1007/BF00992696}
}

@misc{schulman2017ppo,
  title         = {Proximal Policy Optimization Algorithms},
  author        = {Schulman, John and Wolski, Filip and Dhariwal, Prafulla and Radford, Alec and Klimov, Oleg},
  year          = {2017},
  eprint        = {1707.06347},
  archivePrefix = {arXiv},
  primaryClass  = {cs.LG},
  url           = {https://arxiv.org/abs/1707.06347}
}

@misc{shao2024deepseekmath,
  title         = {DeepSeekMath: Pushing the Limits of Mathematical Reasoning in Open Language Models},
  author        = {Shao, Zhihong and Wang, Peiyi and Zhu, Qihao and Xu, Runxin and Song, Junxiao and Bi, Xiao and Zhang, Haowei and Zhang, Mingchuan and Li, Y. K. and Wu, Y. and Guo, Daya},
  year          = {2024},
  eprint        = {2402.03300},
  archivePrefix = {arXiv},
  primaryClass  = {cs.CL},
  doi           = {10.48550/arXiv.2402.03300},
  url           = {https://arxiv.org/abs/2402.03300}
}

@misc{espeholt2018impala,
  title         = {IMPALA: Scalable Distributed Deep-RL with Importance Weighted Actor-Learner Architectures},
  author        = {Espeholt, Lasse and Soyer, Hubert and Munos, R{\'e}mi and Simonyan, Karen and Mnih, Volodyrmyr and Ward, Tom and Doron, Yotam and Firoiu, Vlad and Harley, Tim and Dunning, Iain and Legg, Shane and Kavukcuoglu, Koray},
  year          = {2018},
  eprint        = {1802.01561},
  archivePrefix = {arXiv},
  primaryClass  = {cs.LG},
  url           = {https://arxiv.org/abs/1802.01561}
}

@misc{fu2025areal,
  title         = {AReaL: A Large-Scale Asynchronous Reinforcement Learning System for Language Reasoning},
  author        = {Fu, Wei and Gao, Jiaxuan and Shen, Xujie and Zhu, Chen and Mei, Zhiyu and He, Chuyi and Xu, Shusheng and Wei, Guo and Mei, Jun and Wang, Jiashu and Yang, Tongkai and Yuan, Binhang and Wu, Yi},
  year          = {2025},
  eprint        = {2505.24298},
  archivePrefix = {arXiv},
  primaryClass  = {cs.LG},
  url           = {https://arxiv.org/abs/2505.24298}
}

@misc{wu2025llamarl,
  title         = {LlamaRL: A Distributed Asynchronous Reinforcement Learning Framework for Efficient Large-scale LLM Training},
  author        = {Wu, Bo and Wang, Sid and Tang, Yunhao and Ding, Jia and Helenowski, Eryk and Tan, Liang and Xu, Tengyu and Gowda, Tushar and Chen, Zhengxing and Zhu, Chen and Tang, Xiaocheng and Qian, Yundi and Zhu, Beibei and Hou, Rui},
  year          = {2025},
  eprint        = {2505.24034},
  archivePrefix = {arXiv},
  primaryClass  = {cs.LG},
  url           = {https://arxiv.org/abs/2505.24034}
}

@misc{piche2025pipelinerl,
  title         = {PipelineRL: Faster On-policy Reinforcement Learning for Long Sequence Generation},
  author        = {Pich{\'e}, Alexandre and Kamaloo, Ehsan and Pardinas, Rafael and Bahdanau, Dzmitry},
  year          = {2025},
  eprint        = {2509.19128},
  archivePrefix = {arXiv},
  primaryClass  = {cs.LG},
  url           = {https://arxiv.org/abs/2509.19128}
}

@misc{hu2025taminglongtail,
  title         = {Taming the Long-Tail: Efficient Reasoning RL Training with Adaptive Drafter},
  author        = {Hu, Qinghao and Yang, Shang and Guo, Junxian and Yao, Xiaozhe and Lin, Yujun and Gu, Yuxian and Cai, Han and Gan, Chuang and Klimovic, Ana and Han, Song},
  year          = {2025},
  eprint        = {2511.16665},
  archivePrefix = {arXiv},
  primaryClass  = {cs.LG},
  url           = {https://arxiv.org/abs/2511.16665}
}

@book{kish1995surveysampling_reprint,
  title     = {Survey Sampling},
  author    = {Kish, Leslie},
  publisher = {John Wiley \& Sons},
  year      = {1995},
  isbn      = {9780471109495},
  note      = {Later reprint/edition listing; original publication 1965.}
}

@techreport{kong1992standardized,
  title       = {A Note on Importance Sampling using Standardized Weights},
  author      = {Kong, Augustine},
  institution = {Department of Statistics, The University of Chicago},
  type        = {Technical Report},
  number      = {348},
  year        = {1992},
  month       = jul,
  url         = {https://d3qi0qp55mx5f5.cloudfront.net/stat/docs/tech-rpts/tr348.pdf}
}

@article{kong1994sequential,
  title   = {Sequential Imputations and {B}ayesian Missing Data Problems},
  author  = {Kong, Augustine and Liu, Jun S. and Wong, Wing Hung},
  journal = {Journal of the American Statistical Association},
  volume  = {89},
  number  = {425},
  pages   = {278--288},
  year    = {1994},
  doi     = {10.1080/01621459.1994.10476469},
  url     = {https://www.tandfonline.com/doi/abs/10.1080/01621459.1994.10476469}
}

@misc{goyal2017largemini,
  title         = {Accurate, Large Minibatch {SGD}: Training {ImageNet} in 1 Hour},
  author        = {Goyal, Priya and Doll{\'a}r, Piotr and Girshick, Ross and Noordhuis, Pieter and Wesolowski, Lukasz and Kyrola, Aapo and Tulloch, Andrew and Jia, Yangqing and He, Kaiming},
  year          = {2017},
  eprint        = {1706.02677},
  archivePrefix = {arXiv},
  primaryClass  = {cs.CV},
  url           = {https://arxiv.org/abs/1706.02677}
}

@incollection{dayan1991reinforcementcomparison,
  title     = {Reinforcement Comparison},
  author    = {Dayan, Peter},
  booktitle = {Connectionist Models: Proceedings of the 1990 Summer School},
  editor    = {Touretzky, David S. and Elman, Jeffrey L. and Sejnowski, Terrence J. and Hinton, Geoffrey E.},
  pages     = {45--51},
  year      = {1991},
  publisher = {Morgan Kaufmann},
  address   = {San Mateo, CA}
}

@article{greensmith2004variance,
  title   = {Variance Reduction Techniques for Gradient Estimates in Reinforcement Learning},
  author  = {Greensmith, Evan and Bartlett, Peter L. and Baxter, Jonathan},
  journal = {Journal of Machine Learning Research},
  volume  = {5},
  number  = {Nov},
  pages   = {1471--1530},
  year    = {2004},
  url     = {https://www.jmlr.org/papers/v5/greensmith04a.html}
}

@misc{weaver2013optimalbaseline,
  title         = {The Optimal Reward Baseline for Gradient-Based Reinforcement Learning},
  author        = {Weaver, Lex and Tao, Nigel},
  year          = {2013},
  eprint        = {1301.2315},
  archivePrefix = {arXiv},
  primaryClass  = {cs.LG},
  url           = {https://arxiv.org/abs/1301.2315}
}

@misc{li2023remax,
  title         = {ReMax: A Simple, Effective, and Efficient Reinforcement Learning Method for Aligning Large Language Models},
  author        = {Li, Ziniu and Xu, Tian and Zhang, Yushun and Lin, Zhihang and Yu, Yang and Sun, Ruoyu and Luo, Zhi-Quan},
  year          = {2023},
  eprint        = {2310.10505},
  archivePrefix = {arXiv},
  primaryClass  = {cs.CL},
  url           = {https://arxiv.org/abs/2310.10505}
}

@misc{hao2025opo,
  title         = {On-Policy RL with Optimal Reward Baseline},
  author        = {Hao, Yaru and Dong, Li and Wu, Xun and Huang, Shaohan and Chi, Zewen and Wei, Furu},
  year          = {2025},
  eprint        = {2505.23585},
  archivePrefix = {arXiv},
  primaryClass  = {cs.LG},
  url           = {https://arxiv.org/abs/2505.23585}
}

@inproceedings{sutton2000policygradient,
  title     = {Policy Gradient Methods for Reinforcement Learning with Function Approximation},
  author    = {Sutton, Richard S. and McAllester, David and Singh, Satinder and Mansour, Yishay},
  booktitle = {Advances in Neural Information Processing Systems (NeurIPS)},
  year      = {2000},
  url       = {https://papers.nips.cc/paper/1713-policy-gradient-methods-for-reinforcement-learning-with-function-approximation}
}

@misc{degris2012offpolicyactorcritic,
  title         = {Off-Policy Actor-Critic},
  author        = {Degris, Thomas and White, Martha and Sutton, Richard S.},
  year          = {2012},
  eprint        = {1205.4839},
  archivePrefix = {arXiv},
  primaryClass  = {cs.LG},
  url           = {https://arxiv.org/abs/1205.4839}
}

@online{he2025defeating_nondeterminism_llm_inference,
  title  = {Defeating Nondeterminism in LLM Inference},
  author = {He, Horace},
  year   = {2025},
  month  = sep,
  url    = {https://thinkingmachines.ai/blog/defeating-nondeterminism-in-llm-inference/},
  note   = {Thinking Machines Lab blog post. Published Sep 10, 2025.}
}

@misc{xi2026jetrl,
  title         = {Jet-RL: Enabling On-Policy FP8 Reinforcement Learning with Unified Training and Rollout Precision Flow},
  author        = {Xi, Haocheng and Ruan, Charlie and Liao, Peiyuan and Lin, Yujun and Cai, Han and Zhao, Yilong and Yang, Shuo and Keutzer, Kurt and Han, Song and Zhu, Ligeng},
  year          = {2026},
  eprint        = {2601.14243},
  archivePrefix = {arXiv},
  primaryClass  = {cs.LG},
  url           = {https://arxiv.org/abs/2601.14243}
}

@misc{cobbe2021gsm8k,
  title         = {Training Verifiers to Solve Math Word Problems},
  author        = {Cobbe, Karl and Kosaraju, Vineet and Bavarian, Mohammad and Chen, Mark and Jun, Heewoo and Kaiser, Lukasz and Plappert, Matthias and Tworek, Jerry and Hilton, Jacob and Nakano, Reiichiro and Hesse, Christopher and Schulman, John},
  year          = {2021},
  eprint        = {2110.14168},
  archivePrefix = {arXiv},
  primaryClass  = {cs.LG},
  url           = {https://arxiv.org/abs/2110.14168}
}

@misc{hendrycks2021math,
  title         = {Measuring Mathematical Problem Solving With the {MATH} Dataset},
  author        = {Hendrycks, Dan and Burns, Collin and Kadavath, Saurav and Arora, Akul and Basart, Steven and Tang, Eric and Song, Dawn and Steinhardt, Jacob},
  year          = {2021},
  eprint        = {2103.03874},
  archivePrefix = {arXiv},
  primaryClass  = {cs.LG},
  url           = {https://arxiv.org/abs/2103.03874}
}

@misc{stojanovski2025reasoninggym,
  title         = {REASONING GYM: Reasoning Environments for Reinforcement Learning with Verifiable Rewards},
  author        = {Stojanovski, Zafir and Stanley, Oliver and Sharratt, Joe and Jones, Richard and Adefioye, Abdulhakeem and Kaddour, Jean and K{\"o}pf, Andreas},
  year          = {2025},
  eprint        = {2505.24760},
  archivePrefix = {arXiv},
  primaryClass  = {cs.LG},
  url           = {https://arxiv.org/abs/2505.24760}
}

@misc{xue2025simpletir,
  title         = {SimpleTIR: End-to-End Reinforcement Learning for Multi-Turn Tool-Integrated Reasoning},
  author        = {Xue, Zhenghai and Zheng, Longtao and Liu, Qian and Li, Yingru and Zheng, Xiaosen and Ma, Zejun and An, Bo},
  year          = {2025},
  eprint        = {2509.02479},
  archivePrefix = {arXiv},
  primaryClass  = {cs.LG},
  url           = {https://arxiv.org/abs/2509.02479}
}

@misc{yu2025dapo,
  title         = {DAPO: An Open-Source LLM Reinforcement Learning System at Scale},
  author        = {Yu, Qiying and others},
  year          = {2025},
  eprint        = {2503.14476},
  archivePrefix = {arXiv},
  primaryClass  = {cs.LG},
  url           = {https://arxiv.org/abs/2503.14476}
}

@misc{liu2025prorl,
  title         = {ProRL: Prolonged Reinforcement Learning Expands Reasoning Boundaries in Large Language Models},
  author        = {Liu, Mingjie and Diao, Shizhe and Lu, Ximing and Hu, Jian and Dong, Xin and Choi, Yejin and Kautz, Jan and Dong, Yi},
  year          = {2025},
  eprint        = {2505.24864},
  archivePrefix = {arXiv},
  primaryClass  = {cs.LG},
  url           = {https://arxiv.org/abs/2505.24864}
}

@article{krizhevsky2014oneweirdtrick,
  title   = {One weird trick for parallelizing convolutional neural networks},
  author  = {Krizhevsky, Alex},
  journal = {arXiv preprint arXiv:1404.5997},
  year    = {2014}
}

@inproceedings{hoffer2017trainlonger,
  title     = {Train longer, generalize better: closing the generalization gap in large batch training of neural networks},
  author    = {Hoffer, Elad and Hubara, Itay and Soudry, Daniel},
  booktitle = {Advances in Neural Information Processing Systems (NeurIPS)},
  year      = {2017},
  eprint    = {1705.08741},
  archivePrefix = {arXiv}
}

@inproceedings{you2020lamb,
  title     = {Large Batch Optimization for Deep Learning: Training BERT in 76 minutes},
  author    = {You, Yang and Li, Jing and Reddi, Sashank and Hseu, Jonathan and Kumar, Sanjiv and Bhojanapalli, Srinadh and Song, Xiaodan and Demmel, James and Keutzer, Kurt and Hsieh, Cho-Jui},
  booktitle = {International Conference on Learning Representations (ICLR)},
  year      = {2020},
  eprint    = {1904.00962},
  archivePrefix = {arXiv}
}

@inproceedings{liu2018breaking,
  title     = {Breaking the Curse of Horizon: Infinite-Horizon Off-Policy Estimation},
  author    = {Liu, Qiang and Li, Lihong and Tang, Ziyang and Zhou, Dengyong},
  booktitle = {Advances in Neural Information Processing Systems (NeurIPS)},
  year      = {2018}
}

@inproceedings{liu2020understanding,
  title     = {Understanding the Curse of Horizon in Off-Policy Evaluation via Conditional Importance Sampling},
  author    = {Liu, Yao and Bacon, Pierre-Luc and Brunskill, Emma},
  booktitle = {Proceedings of the 37th International Conference on Machine Learning (ICML)},
  series    = {Proceedings of Machine Learning Research},
  volume    = {119},
  pages     = {6184--6193},
  year      = {2020},
  publisher = {PMLR}
}

@misc{zheng2025m2po,
  title         = {Prosperity before Collapse: How Far Can Off-Policy RL Reach with Stale Data on LLMs?},
  author        = {Zheng, Haizhong and Zhao, Jiawei and Chen, Beidi},
  year          = {2025},
  eprint        = {2510.01161},
  archivePrefix = {arXiv},
  primaryClass  = {cs.LG},
  url           = {https://arxiv.org/abs/2510.01161}
}

@misc{zheng2025gspo,
  title         = {Group Sequence Policy Optimization},
  author        = {Zheng, Chujie and Liu, Shixuan and Li, Mingze and Chen, Xiong-Hui and Yu, Bowen and Gao, Chang and Dang, Kai and Liu, Yuqiong and Men, Rui and Yang, An and Zhou, Jingren and Lin, Junyang},
  year          = {2025},
  eprint        = {2507.18071},
  archivePrefix = {arXiv},
  primaryClass  = {cs.LG},
  url           = {https://arxiv.org/abs/2507.18071}
}

@misc{qi2025fp16mismatch,
  title         = {Defeating the Training-Inference Mismatch via FP16},
  author        = {Qi, Penghui and Liu, Zichen and Zhou, Xiangxin and Pang, Tianyu and Du, Chao and Lee, Wee Sun and Lin, Min},
  year          = {2025},
  eprint        = {2510.26788},
  archivePrefix = {arXiv},
  primaryClass  = {cs.LG},
  url           = {https://arxiv.org/abs/2510.26788}
}

@misc{deepseekai2025v32,
  title         = {DeepSeek-V3.2: Pushing the Frontier of Open Large Language Models},
  author        = {{DeepSeek-AI} and Liu, Aixin and Mei, Aoxue and Lin, Bangcai and Xue, Bing and Wang, Bingxuan and Xu, Bingzheng and others},
  year          = {2025},
  eprint        = {2512.02556},
  archivePrefix = {arXiv},
  primaryClass  = {cs.CL},
  url           = {https://arxiv.org/abs/2512.02556}
}

@online{liu-li-2025-rl-collapse,
  title  = {When Speed Kills Stability: Demystifying {RL} Collapse from the Training-Inference Mismatch},
  author = {Liu, Jiacai and Li, Yingru and Fu, Yuqian and Wang, Jiawei and Liu, Qian and Jiang, Zhuo},
  year   = {2025},
  month  = sep,
  url    = {https://yingru.notion.site/When-Speed-Kills-Stability-Demystifying-RL-Collapse-from-the-Training-Inference-Mismatch-271211a558b7808d8b12d403fd15edda},
  note   = {Online report / blog post. First published Sep 17, 2025. Accessed 2026-01-14}
}

@online{yao2025offpolicyrlblog,
  title  = {Your Efficient RL Framework Secretly Brings You Off-Policy RL Training},
  author = {Yao, Feng and Liu, Liyuan and Zhang, Dinghuai and Dong, Chengyu and Shang, Jingbo and Gao, Jianfeng},
  year   = {2025},
  month  = aug,
  url    = {https://fengyao.notion.site/off-policy-rl},
  note   = {Online report / blog post. First published Aug 5, 2025; last updated Oct 13, 2025. Accessed 2026-01-14}
}

@inproceedings{yao2025rollouttrainingmismatch,
  title     = {On the Rollout-Training Mismatch in Modern RL Systems},
  author    = {Yao, Feng and Liu, Liyuan and Zhang, Dinghuai and Dong, Chengyu and Shang, Jingbo and Gao, Jianfeng},
  booktitle = {OPT2025: 17th Annual Workshop on Optimization for Machine Learning},
  year      = {2025},
  url       = {https://opt-ml.org/papers/2025/paper116.pdf},
  note      = {Workshop paper. Accessed 2026-01-14}
}

@misc{ma2025r3,
  title         = {Stabilizing MoE Reinforcement Learning by Aligning Training and Inference Routers},
  author        = {Ma, Wenhan and Zhang, Hailin and Zhao, Liang and Song, Yifan and Wang, Yudong and Sui, Zhifang and Luo, Fuli},
  year          = {2025},
  eprint        = {2510.11370},
  archivePrefix = {arXiv},
  primaryClass  = {cs.LG},
  url           = {https://arxiv.org/abs/2510.11370}
}

@online{zheng2025m2po_notion,
  title  = {How Far Can Off-Policy RL Reach with Stale Data on LLMs? (M2PO page)},
  author = {Zheng, Haizhong and Zhao, Jiawei and Chen, Beidi},
  year   = {2025},
  url    = {https://m2po.notion.site/rl-stale-m2po},
  note   = {Online project page (Notion). Accessed 2026-01-14}
}

@online{zhao2025icepop,
  title  = {Small Leak Can Sink a Great Ship---Boost RL Training on MoE with IcePop!},
  author = {Zhao, Xin and Liu, Yongkang and Xu, Kuan and Guo, Jia and Wang, Zihao and Sun, Yan and Kong, Xinyu and Cao, Qianggang and Jiang, Liang and Wen, Zujie and Zhang, Zhiqiang and Zhou, Jun},
  year   = {2025},
  month  = sep,
  url    = {https://ringtech.notion.site/icepop},
  note   = {Online report (Notion). Accessed 2026-01-14}
}

@online{li2025otb,
  title  = {The Optimal Token Baseline: Variance Reduction for Long-Horizon LLM-RL},
  author = {Li, Yingru and Xu, Jiawei and Li, Ziniu and Liu, Jiacai and Liu, Wei and others},
  year   = {2025},
  month  = dec,
  url    = {https://www.notion.so/399211a558b782cfa936014c0d42dfb8?pvs=21},
  note   = {Online report (Notion). Accessed 2026-01-14}
}

@misc{noukhovitch2024asyncrlhf,
  title         = {Asynchronous RLHF: Faster and More Efficient Off-Policy RL for Language Models},
  author        = {Noukhovitch, Michael and Huang, Shengyi and Xhonneux, Sophie and Hosseini, Arian and Agarwal, Rishabh and Courville, Aaron},
  year          = {2024},
  eprint        = {2410.18252},
  archivePrefix = {arXiv},
  primaryClass  = {cs.LG},
  url           = {https://arxiv.org/abs/2410.18252}
}

@misc{mccandlish2018largebatch,
  title         = {An Empirical Model of Large-Batch Training},
  author        = {McCandlish, Sam and Kaplan, Jared and Amodei, Dario and OpenAI Dota Team},
  year          = {2018},
  eprint        = {1812.06162},
  archivePrefix = {arXiv},
  primaryClass  = {cs.LG},
  url           = {https://arxiv.org/abs/1812.06162}
}

@article{shah2026comedy,
  title        = {A Comedy of Estimators: On KL Regularization in RL Training of LLMs},
  author       = {Shah, Vedant and Obando-Ceron, Johan and Jain, Vineet and Bartoldson, Brian and Kailkhura, Bhavya and Mittal, Sarthak and Berseth, Glen and Castro, Pablo Samuel and Bengio, Yoshua and Malkin, Nikolay and Jain, Moksh and Venkatraman, Siddarth and Courville, Aaron},
  journal      = {arXiv preprint arXiv:2512.21852},
  year         = {2026},
  eprint       = {2512.21852},
  archivePrefix= {arXiv},
  primaryClass = {cs.LG},
  note         = {version 2, 6 Jan 2026}
}
\bibliographystyle{icml2026}

\newpage
\appendix
\newpage
\newpage

\section{Why does the Asynchronous RL collapse?}
\label{app:ablation_collapse_examples}

Across model sizes and tasks, we find the same qualitative pattern: runs that remain stable maintain a healthy ESS ratio, while unstable runs exhibit a pronounced ESS-ratio collapse shortly before KL/gradient explosions and performance collapse. This supports our central diagnosis that asynchronous failure is driven by importance-weight degeneracy and dominated off-policy updates.

\begin{figure}[!htb]
\vskip 0.1in
\centering
\includegraphics[width=\linewidth]{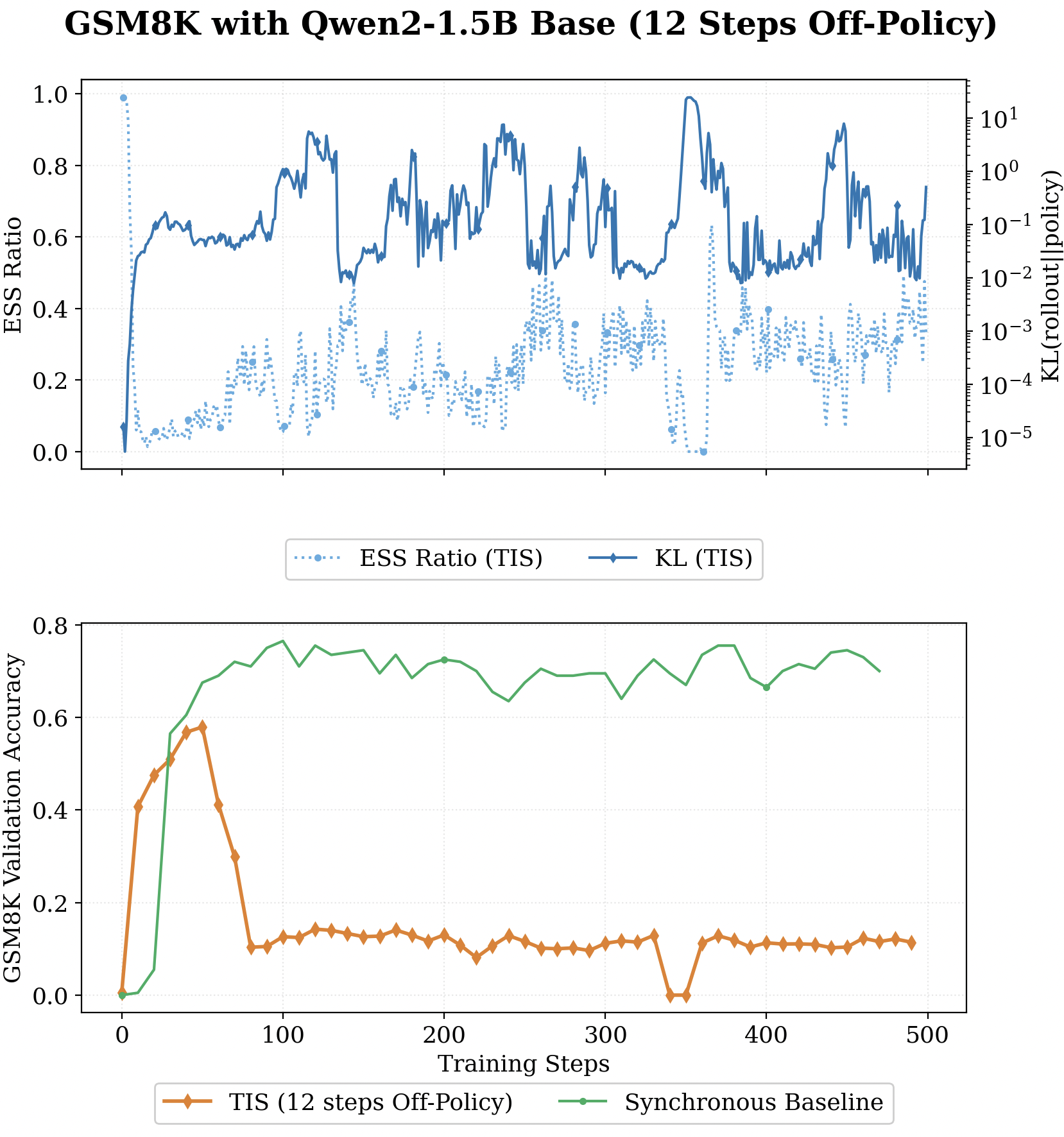}
\vskip -0.05in
\caption{
Qwen2-1.5B Base on GSM8K questions. Truncated Importance Sampling (TIS) exhibits both effective sample size (ESS) collapse and KL explosion, leading to training collapse.
}
\label{fig:gsm8k_collapse}
\end{figure}

\begin{figure}[!htb]
\vskip 0.1in
\centering
\includegraphics[width=\linewidth]{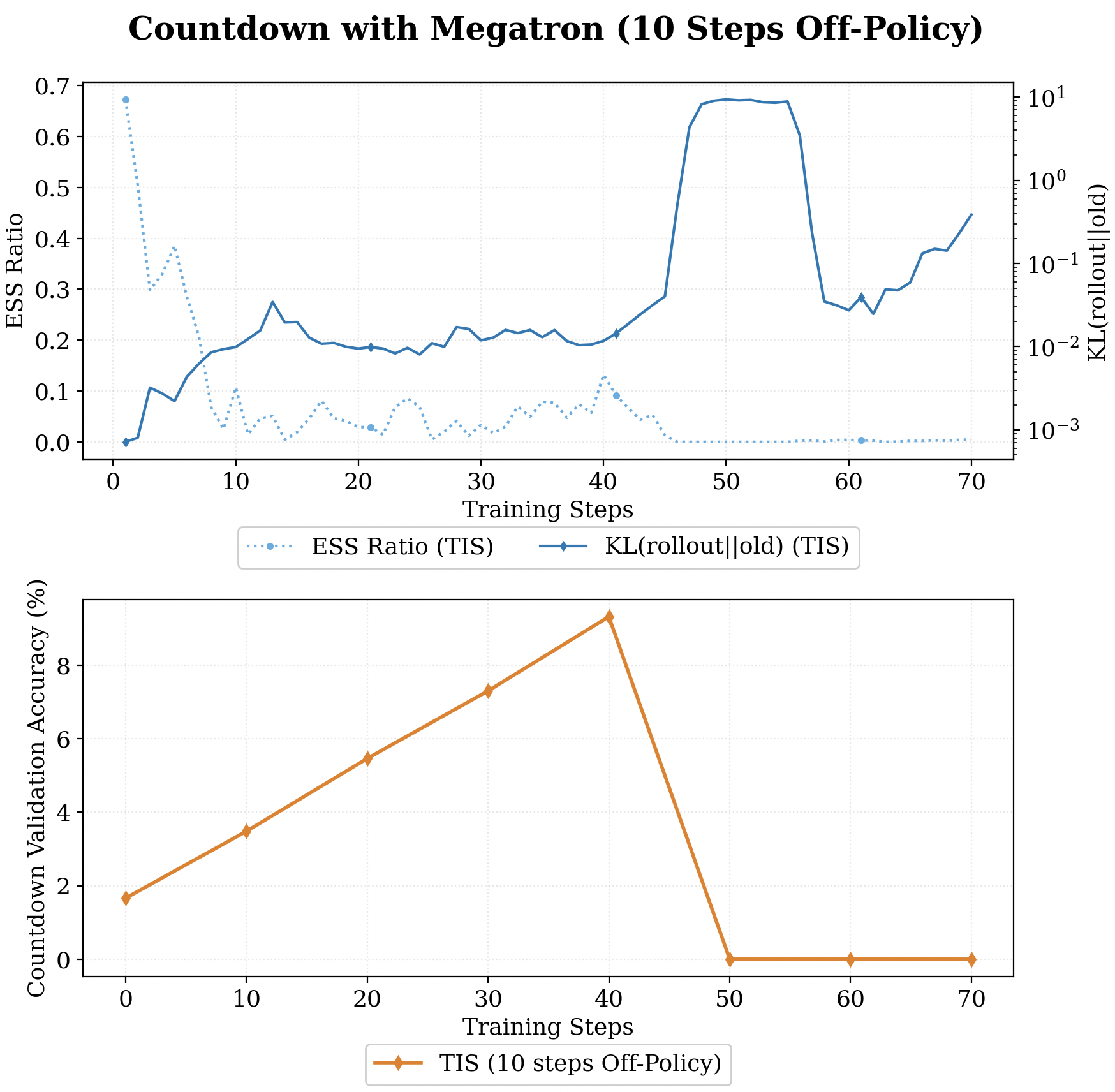}
\vskip -0.05in
\caption{
Countdown (Qwen2-1.5B Base) under high asynchrony with TIS.
Validation accuracy collapses to near zero within the first $\sim$50 training steps, indicating rapid training failure.
}
\label{fig:countdown_collapse}
\end{figure}

In addition to Figure~\ref{fig:didactic}, which illustrates a collapse trajectory on MATH-500, we observe the same qualitative failure pattern across another mathematical reasoning task (GSM8K) and general reasoning task (Countdown). Figure~\ref{fig:gsm8k_collapse} shows a representative GSM8K run under high policy lag ($k{=}12$) using truncated importance sampling (TIS): the ESS ratio degrades and then collapses (i.e., the weighted update becomes dominated by a small subset of trajectories), followed by instability in rollout--policy KL and a sharp drop in task performance. Once ESS collapses, updates become effectively single-sample, making the policy-gradient estimator extremely high variance and precipitating abrupt KL spikes. Similarly, on Countdown (Figure~\ref{fig:countdown_collapse}), validation accuracy collapses to near zero within the first $\sim$50 training steps, consistent with rapid weight degeneracy and dominated off-policy updates.

\section{Off-Policy Optimal Baseline}
\label{app:offpolicy_baseline}

We derive the variance-minimizing \emph{scalar} reward baseline for an importance-weighted policy-gradient estimator.
Throughout, we assume standard regularity conditions for the score-function gradient estimator, including
(i) support: $\mu(\tau)>0$ whenever $\pi_\theta(\tau)>0$ so importance ratios are well-defined, and
(ii) interchange of gradient and expectation so that $\mathbb{E}_{\tau\sim\pi_\theta}[\nabla_\theta \log \pi_\theta(\tau)] = 0$.
These conditions are standard in policy-gradient analyses~\cite{williams1992reinforce,sutton2000policygradient,degris2012offpolicyactorcritic}.

\subsection{Single-sample derivation}
Let a single trajectory sample $\tau$ have reward (or advantage) $R\in\mathbb{R}$ and score-gradient vector
\[
g \;\triangleq\; \nabla_\theta \log \pi_\theta(\tau).
\]
In the off-policy setting, trajectories are sampled from a behavior policy $\mu$, and we use the (unnormalized) importance ratio
\[
w \;\triangleq\; \frac{\pi_\theta(\tau)}{\mu(\tau)}.
\]
Consider the (single-sample) importance-weighted policy-gradient estimator with a constant baseline $b$:
\begin{equation}
G(b) \;=\; w\,(R-b)\,g.
\label{eq:app_single_sample_est}
\end{equation}
Subtracting a constant baseline does not change the expected gradient because
\[
\mathbb{E}_{\tau\sim\mu}[w g]
=
\mathbb{E}_{\tau\sim\pi_\theta}[g]
=
\mathbb{E}_{\tau\sim\pi_\theta}[\nabla_\theta \log \pi_\theta(\tau)]
=0.
\]
Therefore, minimizing $\mathrm{Var}(G(b))$ over $b$ is equivalent to minimizing the second moment
$\mathbb{E}\|G(b)\|_2^2$.

We have
\begin{equation}
\mathbb{E}\|G(b)\|_2^2
=
\mathbb{E}\!\left[w^2 (R-b)^2 \|g\|_2^2\right].
\label{eq:app_second_moment_exact}
\end{equation}
Differentiating~\eqref{eq:app_second_moment_exact} with respect to $b$ yields
\[
\frac{\partial}{\partial b}\,\mathbb{E}\!\left[w^2 (R-b)^2 \|g\|_2^2\right]
=
-2\,\mathbb{E}\!\left[w^2 (R-b)\|g\|_2^2\right].
\]
Setting the derivative to zero gives the variance-minimizing baseline
\begin{equation}
b^\star
=
\frac{\mathbb{E}\!\left[w^2 R \|g\|_2^2\right]}{\mathbb{E}\!\left[w^2 \|g\|_2^2\right]}.
\label{eq:app_offpolicy_opt_baseline_expect}
\end{equation}

\subsection{Minibatch estimator and plug-in form}
For a minibatch of $B$ i.i.d.\ samples $\{(\tau_i,R_i,w_i,g_i)\}_{i=1}^B$ drawn from $\mu$, the minibatch estimator is
\[
\hat G(b) \;=\; \frac{1}{B}\sum_{i=1}^B w_i (R_i-b) g_i.
\]
Since averaging i.i.d.\ samples scales variance by $1/B$, the same $b^\star$ in~\eqref{eq:app_offpolicy_opt_baseline_expect}
minimizes $\mathrm{Var}(\hat G(b))$.

In practice, we use the minibatch plug-in estimate corresponding to~\eqref{eq:app_offpolicy_opt_baseline_expect}:
\begin{equation}
b^\star
\;\approx\;
\frac{\sum_{i=1}^{B} w_i^2 \|g_i\|_2^2 R_i}{\sum_{i=1}^{B} w_i^2 \|g_i\|_2^2}
\label{eq:app_offpolicy_opt_baseline_plugin}
\end{equation}

\paragraph{Remarks.}
(i) In the on-policy regime ($w\equiv 1$), Eq.~\eqref{eq:app_offpolicy_opt_baseline_expect} reduces to the classic gradient-norm-weighted optimal baseline~\cite{greensmith2004variance,weaver2013optimalbaseline}. \\
(ii) Relative to group-mean baselines $b=\frac{1}{B}\sum_i R_i$, Eq.~\eqref{eq:app_offpolicy_opt_baseline_plugin} shows that variance-optimal baselining in the off-policy regime depends on \emph{both} the importance ratios and the gradient magnitudes. In comparison, GRPO adopts the mean normalize and RLOO uses a leave-one-out estimator
\[
b_{\text{GRPO}} = \frac{1}{N} \sum_{i=1}^N R_i, \quad b_{\text{RLOO}}^k = \frac{1}{N - 1}  \sum_{i\neq k} R_i
\]

\section{Efficient Baseline Aware Gradient Computations}
\label{app:opob_implementation}

We implement our baseline-aware updates in Megatron-LM, leveraging its explicit control over parameter shards and optimizer state under DP$\times$TP$\times$CP parallelism.

\paragraph{Per-trajectory gradient-norm statistics under deferred DP synchronization.}
Computing the OPOB baseline requires per-trajectory gradient norms \(\|g_i\|_2\). A naive implementation would perform two backward passes: one to obtain \(\|g_i\|_2\) and a second to form the baseline-adjusted policy gradient. Instead, we compute all required quantities in a \emph{single} backward pass by deferring the data-parallel (DP) gradient all-reduce until after local per-trajectory statistics are computed. Concretely, we run backprop with DP communication disabled so that each worker retains its local gradients, compute the needed per-trajectory scalars locally, and then perform a single DP all-reduce on the final combined gradient.

This design preserves the FLOPs of a single backward pass while replacing the naive two-backward implementation with modest additional memory traffic (for an extra gradient buffer) and a small number of additional scalar reductions.

\paragraph{Two-buffer accumulation for baseline-aware gradients.}
For each trajectory \(\tau_i\), we form the per-trajectory negative log-likelihood
\(\mathcal{L}_i = -\log \pi_\theta(\tau_i)\) and backpropagate once to obtain the score-gradient \(g_i = \nabla_\theta \log \pi_\theta(\tau_i)\).
During this pass, we accumulate \(g_i\) into two gradient buffers:
(i) a reward-weighted buffer with coefficient \(w_i R_i\), and
(ii) a score buffer with coefficient \(w_i\).
After the loop, we compute the OPOB baseline \(b^\star_{\mathrm{OPOB}}\) from the locally accumulated per-trajectory statistics (Algorithm~\ref{alg:single_backward_opob}) and form the final baseline-aware gradient
\(\hat{G} = \frac{1}{B}\left(G_R - b^\star_{\mathrm{OPOB}}\,G_S\right)\)
without any additional forward/backward pass.


\section{Component Ablations of \method}
\label{app:ablation_component}

We ablate \method into its two core variance-control components:  ESS-guided step-size scaling and the off-policy optimal baseline. To quantify each component's contribution under the \emph{same} asynchronous regime and model/task, we compare:
(i) \textbf{TIS} (sequence-level truncated IS only),
(ii) \textbf{TIS + ESS step-size scaling} (our step-size rule applied to AdamW),
(iii) \textbf{TIS + OPOB} (our closed-form off-policy baseline), and
All runs use identical hyperparameters except for the ablated component.

\begin{figure}[!htb]
\vskip 0.1in
\centering
\includegraphics[width=\linewidth]{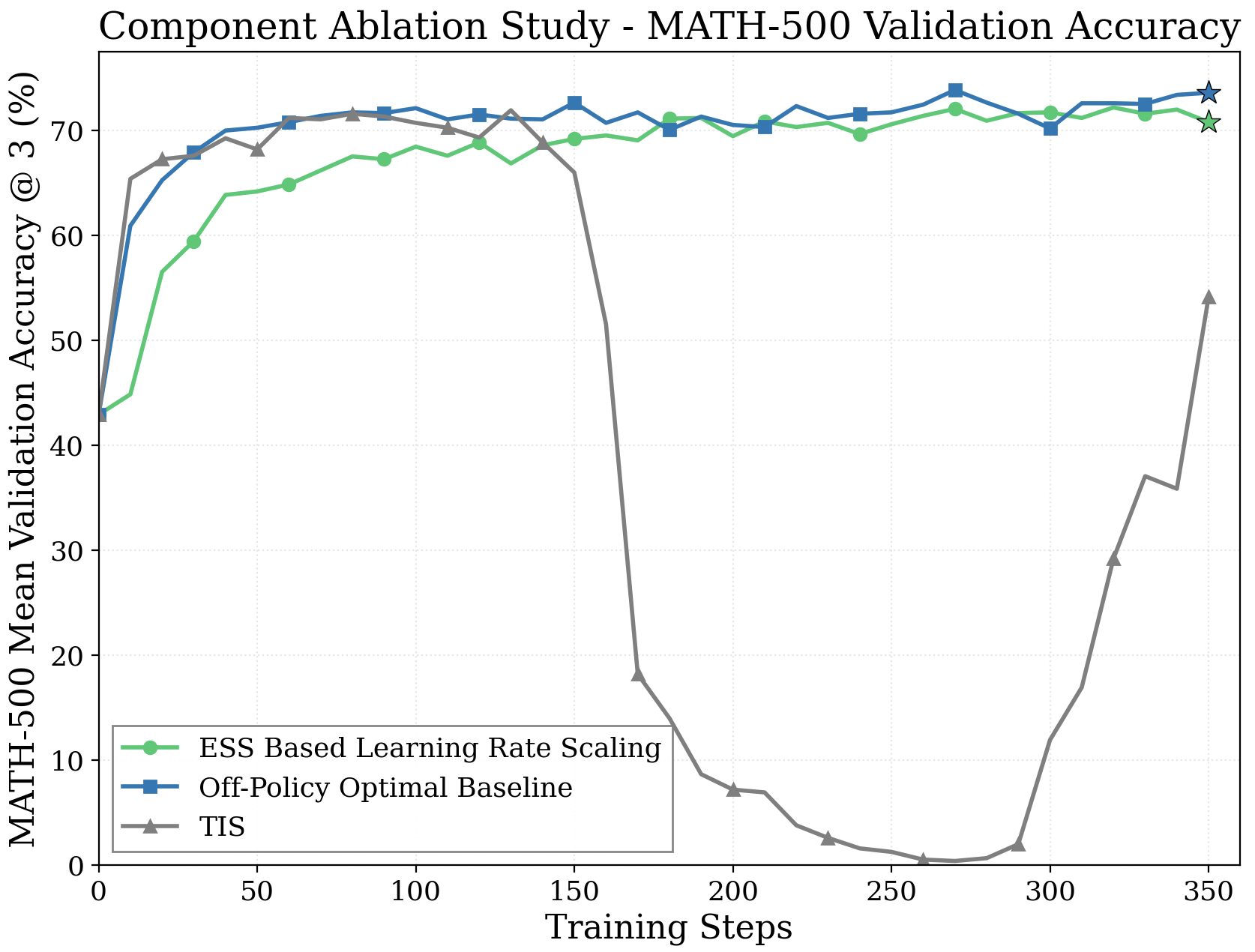}
\vskip -0.05in
\caption{
Component ablations on MATH-500 under high asynchrony (Qwen2.5-7B, \textsc{PipelineRL-}10). We compare TIS alone, as well as adding ESS-guided step-size scaling or the off-policy optimal baseline (OPOB).
Both components improve stability and accuracy.}
\label{fig:component_ablation}
\end{figure}

Figure~\ref{fig:component_ablation} shows that both components improve stability and performance relative to TIS alone, and are complementary.

\section{Further \method Ablation and Comparison Studies}
\label{app:ablation}

This appendix provides additional ablations in the MATH and GSM8K setting to isolate the contributions of (i) the ESS-guided step-size rule and (ii) the optimal off-policy baseline as well compare with previous methods. Unless otherwise noted, we follow the same training setup as in the main experiments Appendix~\ref{app:math_hyperparameters}).

\subsection{Is Lowering the Learning Rate Sufficient for Stable Asynchronous RL?}
\label{app:ablation_lr}

A natural question is whether instability under high asynchrony can be addressed by simply lowering the learning rate. Our ESS-guided rule can be viewed as an \emph{automatic, data-dependent} step-size schedule: it reduces the effective step size precisely when importance weights concentrate, without requiring manual learning-rate sweeps.

\begin{figure}[!htb]
\vskip 0.1in
\centering
\includegraphics[width=\linewidth]{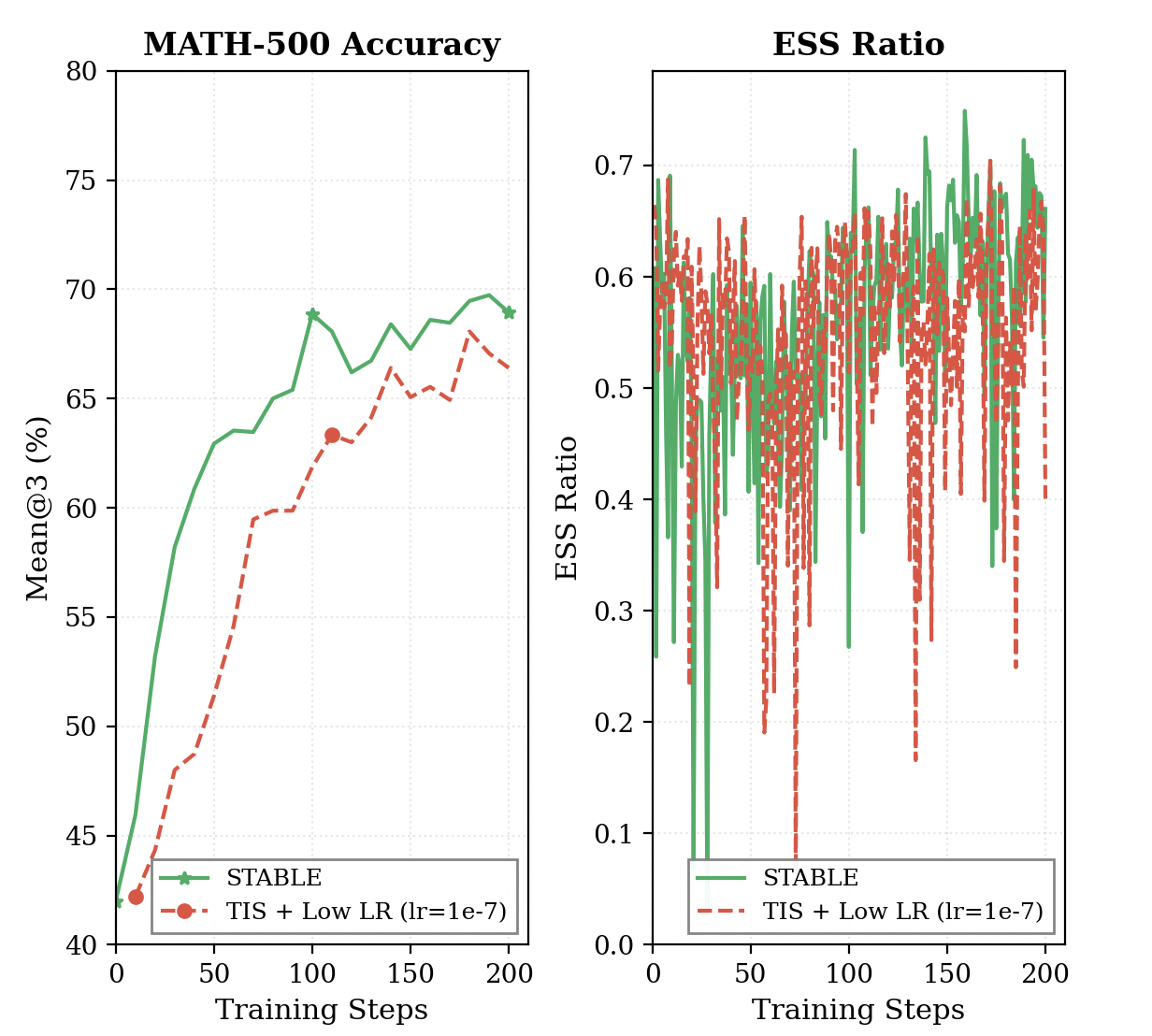}
\vskip -0.05in
\caption{
Lowering the learning rate is not a substitute for variance-aware step-size control.
On MATH-500 under high asynchrony, a very small fixed learning rate (TIS + Low LR) improves stability but converges more slowly and to lower accuracy, despite exhibiting similar ESS levels.
\method reaches higher accuracy with stable updates without learning-rate sweeps.
}
\label{fig:low_lr}
\end{figure}

To test whether conservative tuning alone is sufficient, we run fixed low-learning-rate baselines under the same asynchronous setting. Figure~\ref{fig:low_lr} compares \method to a lower learning rate (with lr=$1^{-7}$ instead of lr=$1^{-6}$) variant of truncated importance sampling (TIS). While reducing the learning rate can mitigate catastrophic collapse, it does so by slowing learning and yielding worse final accuracy. In contrast, \method achieves both higher accuracy and stable training without per-task learning-rate tuning, illustrating a favorable stability--efficiency tradeoff.
 
\subsection{Are Clipping or Masking Methods Sufficient?}
\label{app:ablation_clipping_masking}

We next examine whether standard heuristics that directly bound importance ratios are sufficient to stabilize highly asynchronous RL. We report additional results for both \emph{masking} (MIS-style) and \emph{clipping} (TIS-style) applied to different ratio definitions.

\paragraph{Methods.}
We consider three ratio variants for both masking and clipping:
(i) \textbf{sequence ratio} (full-trajectory likelihood ratio),
(ii) \textbf{token ratio} (per-token likelihood ratio),
and (iii) \textbf{geometric mean ratio} (a length-normalized proxy).
For masking, we zero out the update when the chosen ratio exceeds a threshold; for clipping, we cap the ratio at the threshold.

DeepSeek-v3.2~\cite{deepseekai2025v32} introduces a masking heuristic that is closely related to masked importance sampling, using a \emph{geometric-mean} MIS with an \emph{additional sign constraint on advantages} to discard unstable updates. 

\paragraph{Thresholds.}
Following prior work and practitioner guidance suggesting that bounding extreme ratios can reduce variance~\cite{yao2025offpolicyrlblog,wu2025llamarl,fu2025areal}, we evaluate two representative maximum thresholds, \(c\in\{2.0, 8.0\}\), across all ratio variants. (We keep other hyperparameters matched to the main setup.)

\begin{figure}[!htb]
\vskip 0.1in
\centering
\includegraphics[width=\linewidth]{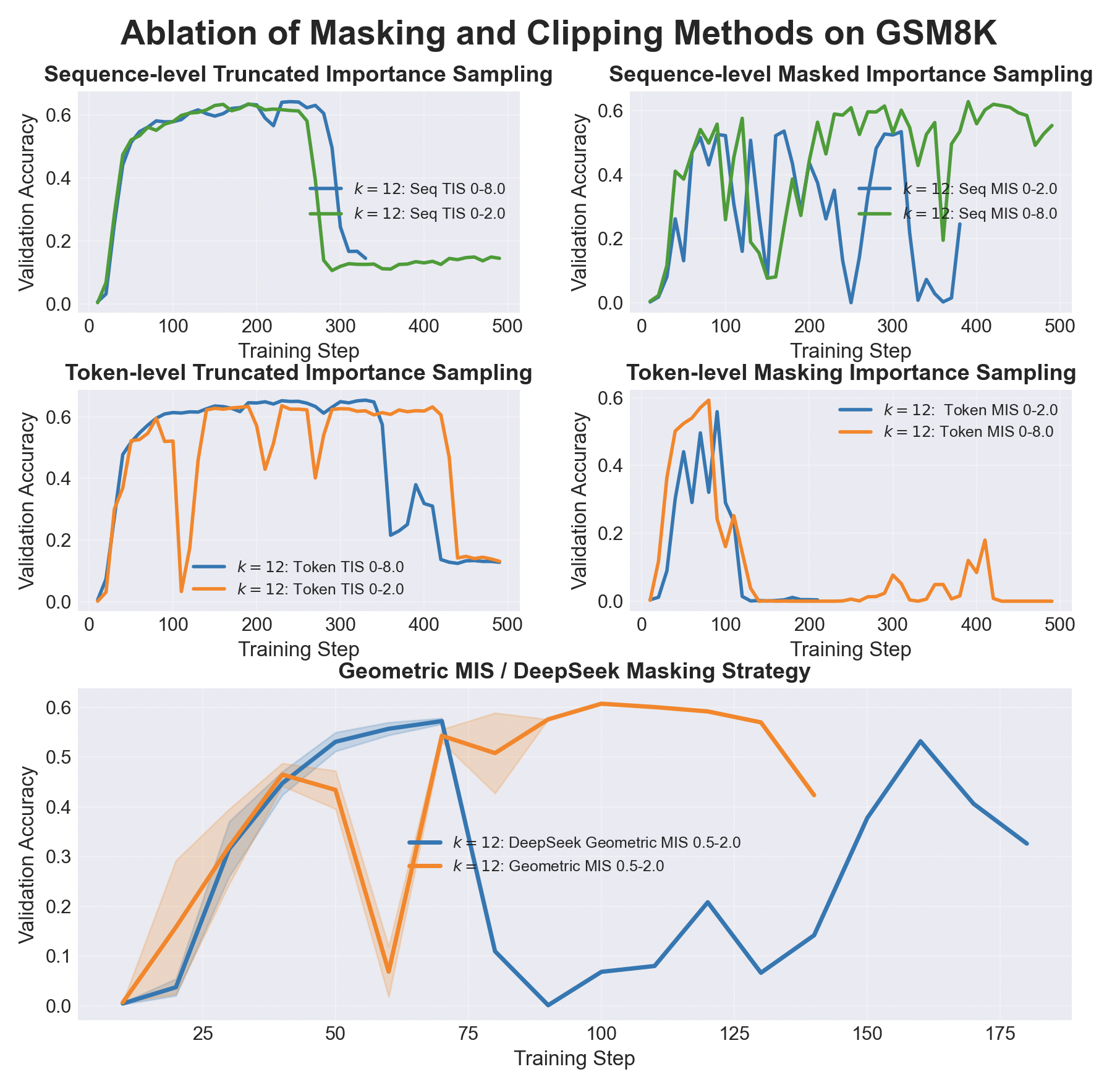}
\vskip -0.05in
\caption{
\textbf{Masking and clipping ablations} under 12 steps off-policy on GSM8K (Qwen2-1.5B). We compare truncated importance sampling (TIS) and masked importance sampling (MIS) using sequence-level, token-level, and geometric-mean ratio statistics with thresholds \(c\in\{2.0,8.0\}\). Overall, these methods bounding is highly sensitive to the threshold and fail to prevent training collapse.
}
\label{fig:masking_clipping}
\end{figure}

In Figure~\ref{fig:masking_clipping}, \textsc{TIS} $(a,b)$ denotes \emph{truncated} importance sampling where ratios outside the interval \((a,b)\) are clipped back into \((a,b)\), while \textsc{MIS} $(a,b)$ denotes \emph{masked} importance sampling where updates with ratios outside \((a,b)\) are discarded (i.e., assigned zero weight).

\paragraph{Findings and tuning sensitivity.}
Figure~\ref{fig:masking_clipping} shows that these masking/clipping methods can delay collapse in some settings, but it is not a reliable substitute for \method under high asynchrony. In particular, sequence-level MIS is highly brittle: at both thresholds we find it can mask nearly all sequences, effectively removing the learning signal and causing training to fail. Token-level variants can be less brittle but remain unstable and sensitive to the threshold choice. Among the baselines we tested, \textit{sequence-level TIS with the larger threshold (\(c=8.0\)) performs best}, but still underperforms \method and does not consistently prevent late-stage instability.

A further limitation is additional overhead of hyperparameter tuning: the importance-ratio distribution becomes broader and more heavy-tailed as model sizes increases, moving to lower preivios, so the stabilizing threshold depends strongly on the degree of off-policy-ness. In practice this requires additional sweeps over both the threshold and the ratio definition (sequence vs.\ token vs.\ geometric mean), which becomes increasingly costly when scaling to new tasks, models, or lag values.

\begin{figure}[!htb]
\vskip 0.1in
\centering
\includegraphics[width=\linewidth]{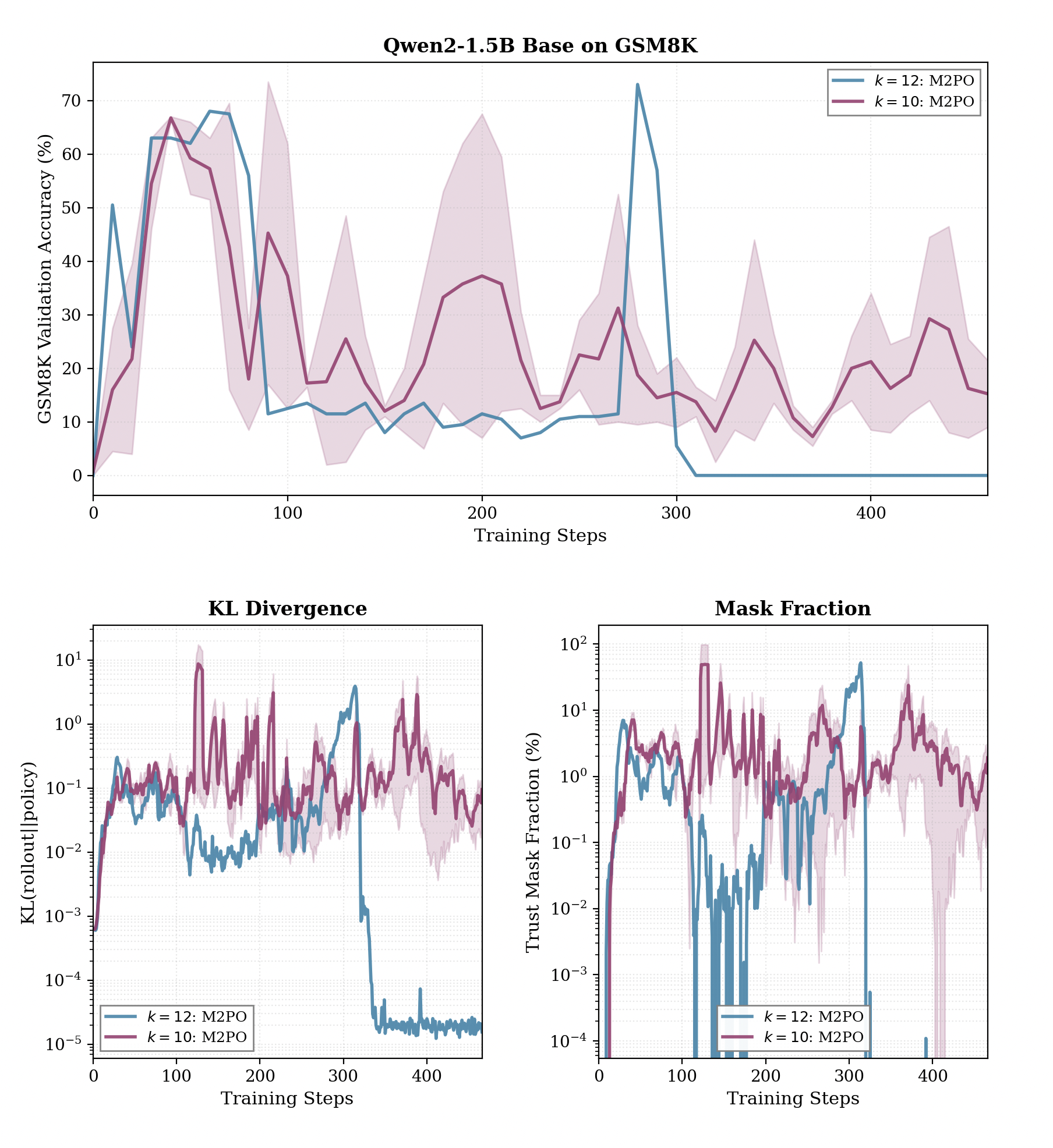}
\vskip -0.05in
\caption{
M2PO under high asynchrony on GSM8K (Qwen2-1.5B). 
Top: GSM8K validation accuracy. Bottom-left: rollout--policy KL divergence (log scale). Bottom-right: trusted-token mask fraction (log scale). 
At $k{=}12$ (and similarly at $k{=}10$), the mask fraction becomes unstable and can collapse toward masking nearly all tokens, coinciding with KL instability and training collapse. Shaded regions indicate run-to-run variability.
}
\label{fig:m2po}
\end{figure}

\paragraph{M2PO-style masking.}
We reproduce \textsc{M2PO}~\cite{zheng2025m2po} using the authors' recommended masking threshold (max $=0.04$). 
Figure~\ref{fig:m2po} shows that at high policy lag (\textsc{PipelineRL}$k{=}12$) on GSM8K with Qwen2-1.5B, \textsc{M2PO} does not yield stable training: the trusted-token (mask) fraction becomes highly erratic and can collapse toward masking nearly all tokens, coinciding with KL spikes and a sharp drop in validation accuracy. We note reducing the lag to $k{=}10$ does not reliably fix this behavior. We suspect that the commonly adopted \textsc{PipelineRL} style asynchronous training setup may differ from the exact regime studied in \textsc{M2PO}. In particular, their pipeline appears to train on exclusively stale rollouts from a fixed-lag behavior policy, whereas our system maintains a queue with \emph{mixed-staleness} samples and may also include in-flight updates with partially completed rollouts. This additional heterogeneity can broaden the importance-ratio distribution and may exacerbate masking instability, so the results in Figure~\ref{fig:m2po} should be interpreted as evaluating M2PO under our (more heterogeneous) \textsc{PipelineRL} style asynchronous stack.

\subsection{Does KL Loss Improve Stability?}
\label{app:ablation_kl}

\begin{figure}[!htb]
\vskip 0.1in
\centering
\includegraphics[width=\linewidth]{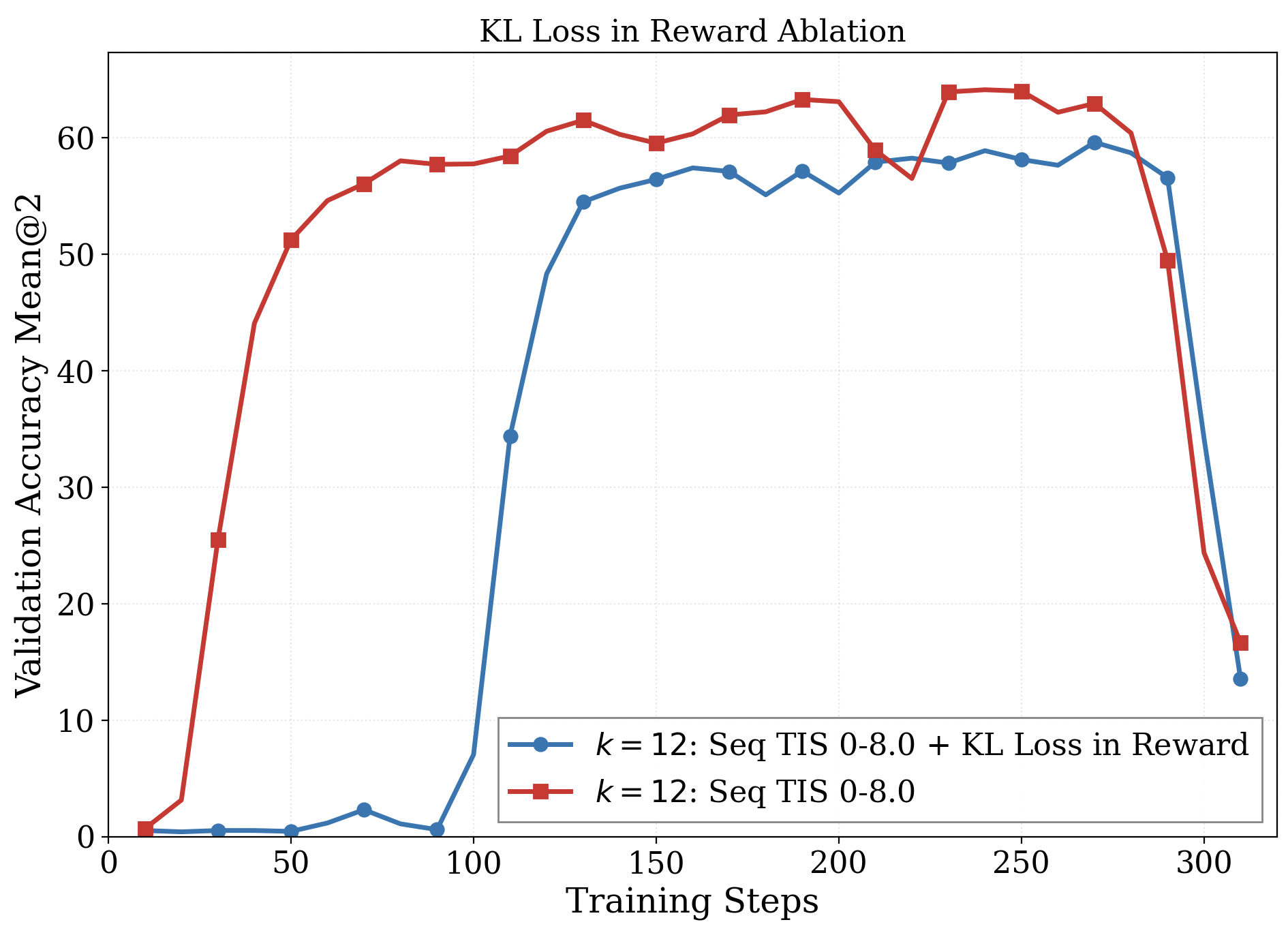}
\vskip -0.05in
\caption{Effect of adding a KL penalty in the reward for \textsc{Seq-TIS} under high asynchrony ($k=12$) on GSM8K with Qwen2-1.5B. KL-in-reward ($\beta=0.001$) does not prevent eventual collapse and also lowers the peak validation accuracy compared to \textsc{Seq-TIS} without KL.}
\label{fig:kl_loss}
\end{figure}

A common stabilization method is to add a KL regularizer ~\cite{schulman2017ppo,shao2024deepseekmath} toward a reference policy, either as an explicit penalty in the objective or by folding it into the reward. To test whether KL regularization alone can mitigate the collapse we observe under large policy lag, we follow prior work~\cite{shah2026comedy} which recommends a KL penalty to the reward (i.e., no direct gradients through the KL term)
\[
R'(x,y) \;=\; R(x,y) \;-\; \beta \,\mathrm{KL}\!\left(\pi_\theta(\cdot \mid x) \,\|\, \pi_{\mathrm{ref}}(\cdot \mid x)\right).
\]
We use a coefficient $\beta = 0.001$ and apply this modification on top of the \textsc{Seq-TIS} baseline in a highly asynchronous setting ($k=12$). As shown in Figure~\ref{fig:kl_loss}, adding KL-in-reward does not prevent collapse: training still eventually becomes unstable and fails. Moreover, KL regularization reduces peak performance, yielding a lower best validation accuracy than \textsc{Seq-TIS} without the KL term. 

\subsection{Are Gradient-Norm Proxies Like Response Length or Logit ``Energy'' Sufficient?}
Two recent works propose replacing per-trajectory gradient norms with cheaper proxies when constructing variance-reducing baselines. For example, OPO uses response length as a proxy for gradient magnitude~\cite{hao2025opo}:
\begin{equation}
\|g_i\| \propto L_i,
\tag{OPO}
\end{equation}
motivated by the intuition that longer responses contribute more log-probability terms.

OTB introduces a logit-based ``energy'' proxy~\cite{li2025otb}. By defining the cumulative per-token \emph{score-function norm}
\begin{equation}
W_t \triangleq \sum_{j=1}^{t}\left\|\nabla_\theta \log \pi_{\theta}(y_j \mid x, y_{<j})\right\|_2^2,
\tag{OTB}
\end{equation}
and approximate the per-token score norm using a closed-form weight
\begin{equation}
\hat{w}_t
=
1 - 2\,\pi_\theta(y_t \mid x, y_{<t}) + \|\boldsymbol{\pi}_t\|_2^2,
\end{equation}
where \(\|\boldsymbol{\pi}_t\|_2^2 = \sum_{v \in \mathcal{V}} \pi_{\theta}^2(v \mid x, y_{<t})\) is the sum of squared next-token probabilities.

While these proxies are attractive for systems reasons, not only are they are derived in the \emph{on-policy} setting, but also in the on-policy setting their approximations weakly correlate with true \(\|g_i\|\).

\begin{figure}[t]
\vskip 0.1in
\centering
\includegraphics[width=\linewidth]{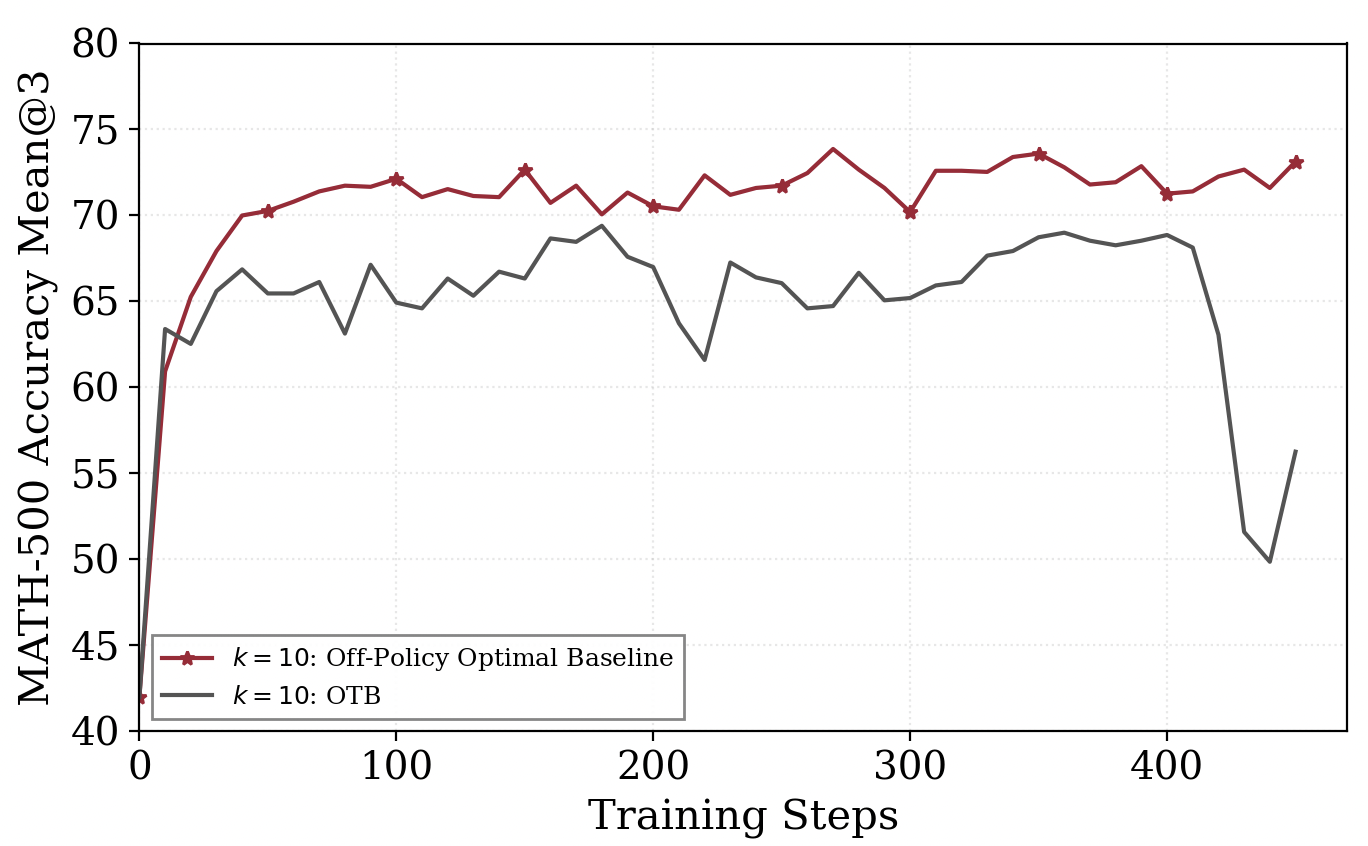}
\vskip -0.05in
\caption{
Qwen2.5-7B Base trained on MATH-500 under high asynchrony ($k{=}10$): the logit-energy proxy baseline (OTB) eventually becomes unstable and collapses, while our off-policy optimal baseline (OPOB) remains stable and sustains higher validation accuracy throughout training.
}

\label{fig:otb_vs_grad_baselining}
\end{figure}

To disentangle the role of our ESS scaling rule, we compare OTB's proxy with only our \method's off policy optimal baseline technique. Empirically, Figure~\ref{fig:otb_vs_grad_baselining} shows that OTB fails to prevent learning collapse, suggesting that both length and logit ``energy'' are both poorly correlated with \(\|g_i\|\), and baselines constructed from these proxies yield weaker stabilization than using the true per-trajectory gradient norms.

\begin{figure}[!htb]
\vskip 0.1in
\centering
\includegraphics[width=\linewidth]{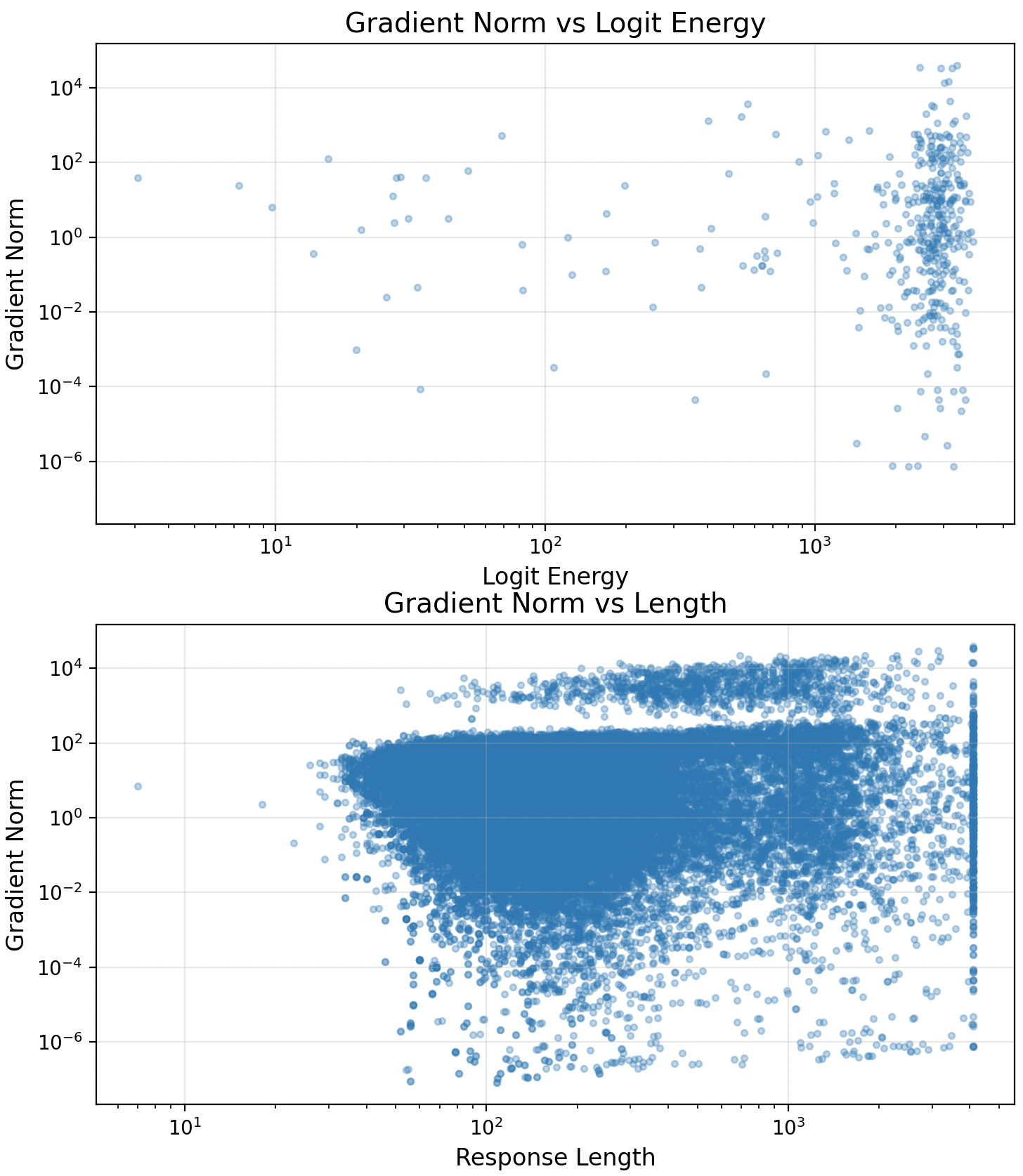}
\vskip -0.05in
\caption{
Qwen2-1.5B Base trajectories on GSM8K questions, with max response length capped to 4096 tokens
}

\label{fig:grad_norm_data}
\end{figure}

Further evidence can be seen in Figure~\ref{fig:grad_norm_data} which 
shows that both OPO and OTB's proxies do not reliably track true per-trajectory gradient norms.

\section{\method Training Details}
\label{app:training_details}

\subsection{Optimization Hyperparameters}
\label{sec:opt_hparams}

We list the hyperparameters for optimization used in all our experiments in Table~\ref{tab:opt_hparams}. 
\begin{table}[!htb]
\centering
\small
\caption{Optimization hyperparameters}
\label{tab:opt_hparams}
\begin{tabular}{lc}
\toprule
\textbf{Hyperparameter} & \textbf{Value} \\
\midrule
Optimizer & AdamW \\
Learning rate & $1\times 10^{-6}$ \\
Warmup steps & 0 \\
Weight decay & 0.1 \\
AdamW $\beta_1,\beta_2$ & 0.9, 0.999 \\
AdamW $\epsilon$ & $10^{-8}$ \\
Gradient clipping & 1.0 \\
Entropy coefficient & 0.0 \\
KL coefficient & 0.0 \\
\bottomrule
\end{tabular}
\end{table}

\subsection{GSM8K}
\label{app:gsm8k_hyperparameters}

\paragraph{Task and reward.}
We train on the GSM8K training split~\cite{cobbe2021gsm8k} using a binary reward that checks whether the model's final numeric answer exactly matches the ground truth. We evaluate on the GSM8K test set and report accuracy (exact-match) over the full set.

\paragraph{Asynchronous setup.}
We use \textsc{PipelineRL-}$k$ asynchronous rollouts with maximum policy lag $k{=}12$, and compute sequence-level importance weights between the sampler and learner policies. We use 6$\times$ A6000 GPUs for sampling and 2$\times$ A6000 GPUs for training. All methods (including our baselines) share the same sampling/training pipeline and differ only in the variance-control and/or masking rules applied to the policy-gradient update.

\paragraph{Note on Hardware.}
Prior work~\cite{liu-li-2025-rl-collapse} reports that Ampere-generation GPUs can exhibit larger sampler--learner numerical discrepancies (e.g., in bf16 log-prob computations) which are not observed on Hopper-generation hardware. This leads to training collapse even for fully on-policy RL without policy-lag.
We intentionally run GSM8K on Ampere (A6000) to stress-test robustness under amplified mismatch in addition to policy lag.
Accordingly, this serves as a worst-case stress test: absolute metrics may change on other hardware, but the relative stability gains of our method over baselines are consistent in our checks.

\begin{table}[!htb]
\centering
\small
\caption{Training hyperparameters for the GSM8K task}
\label{tab:tba_gsm8k_hparams}
\begin{tabularx}{\linewidth}{@{}p{0.46\linewidth}p{0.26\linewidth}X@{}}
\toprule
\textbf{Hyperparameter} & \textbf{Value} & \textbf{Reference} \\
\midrule
Model & Qwen2-1.5B& \\
Learning Rate & $1 \times 10^{-6}$ & \\
Learning Rate Schedule & Warmup Stable Decay & \\
Learning Rate Warmup Steps & 5 & \\
Learning Rate Stable Steps & 395 & \\
Learning Rate Decay Steps & 0 & \\
Generation Temperature & 1.0 & \\
Max Prompt Token Length & 1024 & \\
Response Length & 2048 & \\
Number of Prompts per Batch & 8 & \\
Number of Completions per Prompt & 8 & \\
Batch Size (effective) & 64 & \\
Number of Training Steps & 400 & \\
Total Prompts Seen & 3200 & \\
Total Episodes & 25600 & \\
\midrule
\multicolumn{3}{c}{\textit{Asynchronous-specific hyperparameters}} \\
\midrule
Max Staleness & 12 & \textsc{PipelineRL}-$k$ value \\
Number of Sampler GPUs & 6 & \\
Number of Trainer GPUs & 2 & \\
Max IS ratio & 2.0/8.0 & Max value from TIS \\
$\rho_{\text{ess}}^{\text{on}}$ & 1.0 & On-policy ESS ratio (See \ref{eq:lr_ess_analogy}) \\
\bottomrule
\end{tabularx}
\end{table}

\subsection{Countdown and MATH}
\label{app:math_hyperparameters}

\paragraph{Countdown (Reasoning Gym).}
We build a Countdown-style arithmetic reasoning dataset using Reasoning Gym~\cite{stojanovski2025reasoninggym}.
Rewards are verifiable by deterministic checking of the final answer; we use 9{,}000 problems for training and 1{,}000 held-out problems for validation.

\paragraph{MATH.}
We train on the official MATH training split~\cite{hendrycks2021math} with an exact-match final-answer reward.
We report validation accuracy on the standard MATH validation split, and additionally use the MATH-500 subset for controlled comparisons in the main paper.

\paragraph{Asynchronous setup.}
For both tasks, we run \textsc{PipelineRL-}$k$ with maximum policy lag $k{=}10$. We use $4\times$H100 GPUs for sampling and $4\times$H100 GPUs for training.

\begin{table}[!htb]
\centering
\small
\caption{Training hyperparameters for the MATH and Countdown task}

\begin{tabularx}{\linewidth}{@{}p{0.46\linewidth}p{0.26\linewidth}X@{}}
\toprule
\textbf{Hyperparameter} & \textbf{Value} & \textbf{Reference} \\
\midrule
Model & Qwen2.5-7B & \\
Learning Rate & $1 \times 10^{-6}$ & \\
Learning Rate Schedule & Warmup Stable Decay & \\
Learning Rate Warmup Steps & 0 & \\
Learning Rate Stable Steps & 400 & \\
Learning Rate Decay Steps & 0 & \\
Generation Temperature & 1.0 & \\
Max Prompt Token Length & 2048 & \\
Response Length & 2048 & \\
Number of Prompts per Batch & 32 & \\
Number of Completions per Prompt & 16 & \\
Batch Size (effective) & 512 & \\
Number of Training Steps & 400 & \\
Total Prompts Seen & 12,800 & \\
Total Episodes & 204,800 & \\
\midrule
\multicolumn{3}{c}{\textit{Asynchronous-specific hyperparameters}} \\
\midrule
Max Staleness & 10 & \textsc{PipelineRL}-$k$ value \\
Number of Sampler GPUs & 4 & \\
Number of Trainer GPUs & 4 & \\
Max IS ratio & 8.0 & Max value from TIS \\
$\rho_{\text{ess}}^{\text{on}}$ & 1.0 & On-policy ESS ratio (See \ref{eq:lr_ess_analogy}) \\
\bottomrule
\end{tabularx}
\end{table}

\subsection{Multi-Turn Tool Integrated Reasoning}
\label{app:multiturn_hyperparameters}

\paragraph{Task, data, and evaluation.}
We follow the SimpleTIR setting~\cite{xue2025simpletir} for multi-turn tool-integrated reasoning.
We train on the DAPO-17K dataset~\cite{yu2025dapo} and evaluate on held-out exam benchmarks (AIME 2025), reporting exact-match final-answer accuracy.
We use Qwen2.5-7B Base (rather than newer Qwen3 variants) to reduce the risk of evaluation-set contamination.

\paragraph{Rollout Configuration and Asynchrony.}
We generate up to 5 turns per episode with a maximum completion length of 12{,}288 tokens (Table~\ref{app:multiturn_hyperparameters}).
We run asynchronous training with moderate policy lag (\textsc{PipelineRL-}2) and compute sequence-level importance weights for off-policy correction. We use $2\times$ B200 GPUs for sampling and $2\times$ B200 GPUs for training

\begin{table}[!htb]
\centering
\small
\caption{Training hyperparameters for the multi-turn tool integrated reasoning}
\begin{tabularx}{\linewidth}{@{}p{0.46\linewidth}p{0.26\linewidth}X@{}}
\toprule
\textbf{Hyperparameter} & \textbf{Value} & \textbf{Reference} \\
\midrule
Model & Qwen2.5-7B & \\
Learning Rate & $1 \times 10^{-6}$ & \\
Learning Rate Schedule & Warmup Stable Decay & \\
Learning Rate Warmup Steps & 0 & \\
Learning Rate Stable Steps & 200 & \\
Learning Rate Decay Steps & 0 & \\
Generation Temperature & 1.0 & \\
Max Prompt Token Length & 2048 & \\
Response Length & 12,288 & \\
Number of Prompts per Batch & 128 & \\
Number of Completions per Prompt & 16 & \\
Batch Size (effective) & 2048 & \\
Number of Training Steps & 200 & \\
Total Prompts Seen & 25,600 & \\
Total Episodes & 409,600 & \\
\midrule
\multicolumn{3}{c}{\textit{Asynchronous-specific hyperparameters}} \\
\midrule
Max Staleness & 2 & \textsc{PipelineRL}-$k$ value \\
Number of Sampler GPUs & 2 & \\
Number of Trainer GPUs & 2 & \\
Max IS ratio & 8.0 & Max value from TIS \\
$\rho_{\text{ess}}^{\text{on}}$ & 0.55 & On-policy ESS ratio (See \ref{eq:lr_ess_analogy}) \\
\bottomrule
\end{tabularx}
\end{table}

\end{document}